\documentclass[11pt]{article}
                                    
\usepackage{acl}

\ifx\linenumbers\undefined
  \newcommand{\linenumbers}{}
\fi

\hypersetup{draft}
\newcommand{\clickableurl}[1]{%
  \begingroup\hypersetup{draft=false,colorlinks=true,urlcolor=darkblue}%
  \href{#1}{\textcolor{darkblue}{\nolinkurl{#1}}}\endgroup
}

\usepackage{times}
\usepackage{latexsym}
\usepackage[T1]{fontenc}
\usepackage[utf8]{inputenc}
\usepackage{microtype}
\usepackage{inconsolata}
\usepackage{graphicx}
\usepackage{amssymb}
\usepackage{amsmath}
\usepackage{booktabs}
\usepackage{longtable}
\usepackage{enumitem}
\usepackage{xcolor}
\usepackage{hyperref}
\makeatletter
\let\SA@hyper@@link\hyper@@link
\let\SA@hyper@link@\hyper@link@
\let\SA@hyper@link\hyper@link
\let\SA@hyper@linkurl\hyper@linkurl
\let\SA@hyper@linkstart\hyper@linkstart
\let\SA@hyper@linkend\hyper@linkend
\makeatother
\usepackage{float}
\usepackage{placeins}
\usepackage{comment}
\usepackage[many]{tcolorbox}
\tcbuselibrary{breakable, skins}

\newtcolorbox{promptbox}[2]{%
  enhanced jigsaw, breakable,
  colback=blue!2!white, colframe=blue!35!black, colbacktitle=blue!12!white,
  coltitle=black, fonttitle=\small\bfseries,
  boxrule=0.4pt, titlerule=0pt, arc=2pt,
  left=4pt, right=4pt, top=3pt, bottom=3pt, boxsep=2pt,
  lefttitle=6pt, righttitle=6pt, toptitle=3pt, bottomtitle=3pt,
  before upper={\ttfamily\footnotesize},
  title={#1}, label={#2},
  before skip=0.2cm, after skip=0.2cm,
}

\newtcolorbox{jsonbox}[2]{%
  enhanced jigsaw, breakable,
  colback=blue!2!white, colframe=blue!35!black, colbacktitle=blue!12!white,
  coltitle=black, fonttitle=\small\bfseries,
  boxrule=0.4pt, titlerule=0pt, arc=2pt,
  left=4pt, right=4pt, top=3pt, bottom=3pt, boxsep=2pt,
  lefttitle=6pt, righttitle=6pt, toptitle=3pt, bottomtitle=3pt,
  before upper={\ttfamily\footnotesize},
  title={#1}, label={#2},
  before skip=0.2cm, after skip=0.2cm,
}

\newtcolorbox{reviewbox}[2]{%
  enhanced jigsaw, breakable,
  colback=blue!2!white, colframe=blue!35!black, colbacktitle=blue!12!white,
  coltitle=black, fonttitle=\small\bfseries,
  boxrule=0.4pt, titlerule=0pt, arc=2pt,
  left=5pt, right=5pt, top=3pt, bottom=3pt, boxsep=2pt,
  lefttitle=6pt, righttitle=6pt, toptitle=3pt, bottomtitle=3pt,
  before upper={\small},
  title={#1}, label={#2},
  before skip=0.2cm, after skip=0.2cm,
}

\newcommand{\appendixspacing}{%
  \setlength{\intextsep}{6pt plus 2pt minus 2pt}%
  \setlength{\textfloatsep}{6pt plus 2pt minus 2pt}%
  \setlength{\floatsep}{6pt plus 2pt minus 2pt}%
  \setlength{\abovecaptionskip}{3pt}%
  \setlength{\belowcaptionskip}{0pt}%
}

\title{SA-Bench: Evaluating Semantic Alignment in LLM-Based Paper Reproduction}

\author{
  \textbf{Xue Hu}$^{*}$\textsuperscript{1},
  \textbf{Zewei Pan}$^{*}$\textsuperscript{2},
  \textbf{Zeli Su}\textsuperscript{3},
  \textbf{Zhou Liu}\textsuperscript{4},
  \textbf{Wentao Zhang}$^{\dagger}$\textsuperscript{4,5,6} \\
  \textsuperscript{1}Beihang University, \textsuperscript{2}Shanghai Jiao Tong University, \textsuperscript{3}Minzu University of China \\
  \textsuperscript{4}Peking University, \textsuperscript{5}Zhongguancun Academy \\
  \textsuperscript{6}Beijing Key Laboratory of Data Intelligence and Security (Peking University) \\
  \small{\texttt{kernel@buaa.edu.cn}, \texttt{pzp0057@sjtu.edu.cn},
  \texttt{rickamorty@muc.edu.cn}} \\
  \small{\texttt{zhouliu25@stu.pku.edu.cn}, \texttt{wentao.zhang@pku.edu.cn}} \\
  \small{$^{*}$Equal contribution.\quad $^{\dagger}$Corresponding author.}
}

\begin{document}
\maketitle

\begin{abstract}
LLM agents can generate paper reproduction code, yet often produce
scientifically unfaithful implementations. We define this failure mode as
\textbf{semantic drift}, where generated code silently diverges from the
paper's specifications. We introduce \textsc{SemanticAlign-Bench}
(SA-Bench), a diagnostic benchmark covering 30 papers from ICLR, ICML and
NeurIPS 2025. For each paper, we decompose its specifications into atomic and verifiable implementation claims, which we call \textbf{Semantic Alignment
Units (SAUs)} and evaluate repositories along four diagnostic dimensions
spanning numerical, methodological, protocol and ordering drift. In total,
we construct 1{,}491 SAUs across five ML domains and evaluate 12 generator
configurations (4 models $\times$ 3 scaffolds). Even the strongest
configuration (Claude~+~PaperCoder) achieves a mean SAU score of only 0.301 out of 1.0, with an overall
mean of 0.221 across 360 evaluations. A
failure taxonomy reveals that agents attempt most requirements but implement
them incorrectly, with implementation mismatch and stubs accounting for the
majority of zero-scored claims. Our analysis further indicates that scaffolds
optimized for executability provide limited leverage for scientific
reproduction; narrowing the gap requires scaffolds that prioritize
semantic specification verification. The benchmark, annotations and
evaluation pipeline are publicly available.\footnote{Code:\clickableurl{https://github.com/kernel-14/SemanticAlign-Bench}\\
\hspace*{1.8em}Data:\clickableurl{https://huggingface.co/datasets/kernel-14/SemanticAlign-Bench}}
\end{abstract}

\section{Introduction}
\label{sec:introduction}

\begin{figure*}[t!]
  \centering
  \includegraphics[width=\textwidth]{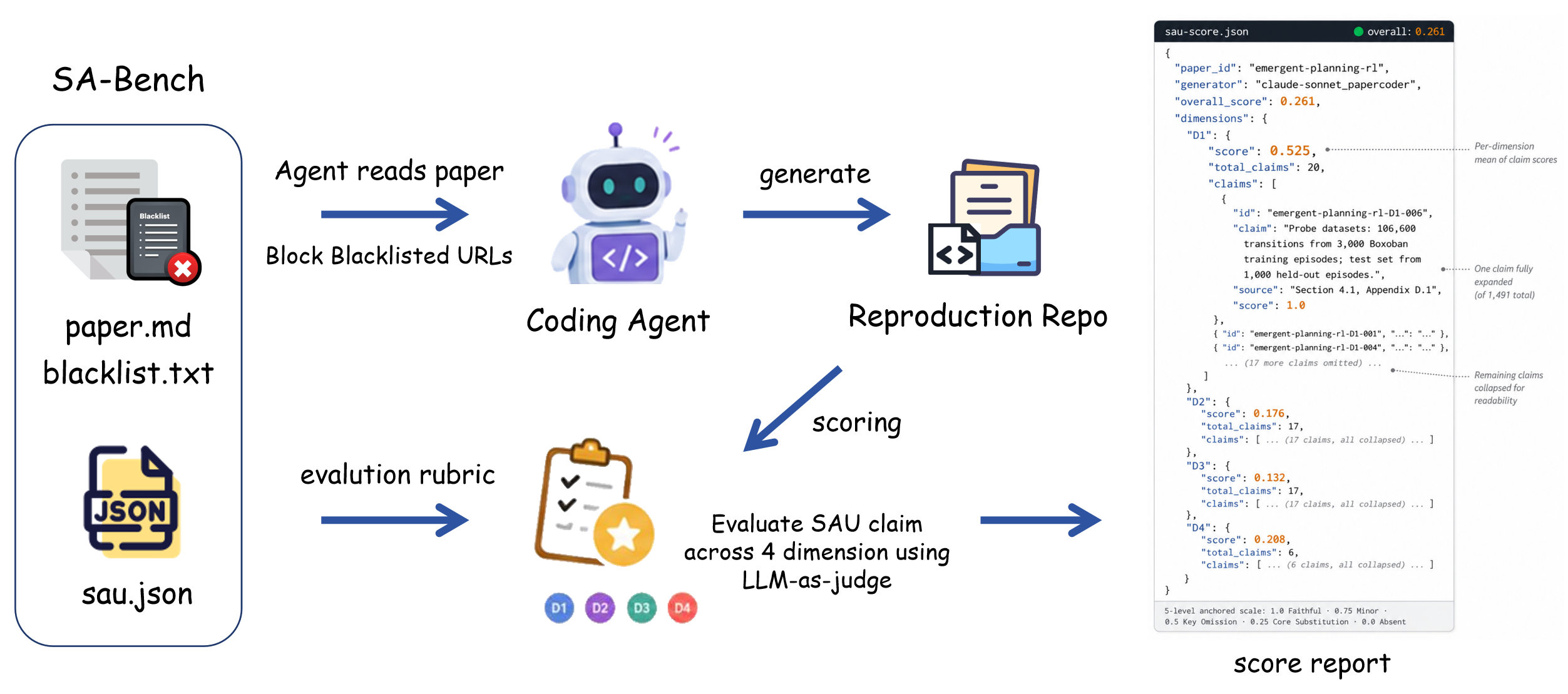}
  \caption{Overview of SA-Bench. The benchmark consists of a curated paper
    collection and per-paper SAU files. Given a paper, a coding agent
    generates a reproduction repository, which is then scored against the
    paper's SAU claims to produce a diagnostic report.
  }
  \label{fig:benchmark-overview}
\end{figure*}

Recent advances in coding agents have propelled their evolution from
single-function completion~\cite{humaneval,mbpp} toward long-horizon
repository generation~\cite{swebench,repobench}. A
representative class of such tasks is
NL2Repo~\cite{nl2repobench}, where agents generate complete repositories from
natural language requirements. Paper-to-code reproduction represents a particularly demanding instance of NL2Repo, as the target repository is expected not merely to be executable, but to faithfully instantiate the scientific specifications of a paper, including its algorithmic logic, numerical details, experimental protocols, and execution order. This shifts the evaluation criterion beyond functional correctness toward semantic fidelity: whether the generated repository implements what the paper actually claims.

This fidelity requirement remains difficult to satisfy for two reasons.
First, scientific papers are written to communicate findings, not to
prescribe implementations: critical details are distributed across
disparate sections, with the core algorithm in the method section and
hyperparameters or preprocessing choices in appendix tables. Second,
scientific code is intolerant of detail-level errors: a single misread
formula or omitted step can fundamentally alter the implemented method,
even if the resulting code remains fully executable. We refer to this
failure mode, where generated code silently diverges from paper
specifications, as \emph{semantic drift}.

Evaluating semantic drift is uniquely challenging because paper reproduction lacks an executable ground-truth oracle. In software engineering benchmarks such as SWE-bench~\cite{swebench}, functional correctness can be verified through reproducible test suites that specify expected program behavior. Scientific reproduction offers no comparable oracle: the authoritative specification is only the paper text itself, and even end-to-end outputs cannot serve as a reliable substitute, since they are confounded by random seeds, software environments, data processing choices and compute budgets. This motivates evaluation methodologies that assess semantic fidelity directly against paper specifications, rather than relying on executability or aggregate empirical outcomes.

Existing benchmarks for paper-to-code reproduction, most notably PaperBench~\cite{paperbench}, evaluate whether generated code reproduces a paper's results, but none offer a systematic classification of the deviation between generated code and paper specifications~\cite{lmrbench,scireplicatebench,scicoqa}. As a result,
semantic drift remains undiagnosed when reproduction falls short.
Table~\ref{tab:benchmark-comparison} summarizes the key differences.

\begin{table}[t]
\centering
\footnotesize
\caption{Comparison between \textsc{SA-Bench} and existing benchmarks. $\checkmark$ denotes supported;
--- denotes not supported.}
\label{tab:benchmark-comparison}
\resizebox{\columnwidth}{!}{%
\begin{tabular}{@{}lccccc@{}}
\toprule
\textbf{Dimension} & \textbf{PaperBench} & \textbf{LMR-BENCH} & \textbf{SciReplicate} & \textbf{SciCoQA} & \textbf{SA-Bench (Ours)} \\
\midrule
Evaluation unit & Rubric node & Function & Function & QA pair & Claim (SAU) \\
Static evaluation & Partial\textsuperscript{*} & --- & --- & \checkmark & \checkmark \\
Failure taxonomy & --- & --- & --- & Partial\textsuperscript{\dag} & \checkmark \\
Graded scoring & Binary & Binary & Binary & Binary & 5-level rubric \\
\bottomrule
\end{tabular}%
}
\vspace{2pt}

{\scriptsize
\textsuperscript{*}PaperBench supports a Code-Development mode use static rubric matching. \\
\textsuperscript{\dag}SciCoQA's categories describe mismatch sources, not
failure diagnostic.
}
\end{table}

To bridge these gaps, we introduce \textsc{SemanticAlign-Bench} (SA-Bench),
a specification-grounded benchmark for diagnosing semantic drift in
paper-to-code reproduction.
%comprising 30 papers spanning ICLR, ICML and NeurIPS~2025 across five ML domains, with 1{,}491 human-verified \emph{Semantic Alignment Units} (SAUs) and 360 paper-level evaluations covering 12 generator configurations (4 models $\times$ 3 scaffolds). 
SA-Bench evaluates \emph{semantic alignment}, the
faithfulness with which generated code implements a paper's verifiable
claims, at the granularity of individual \emph{Semantic Alignment Units}
(SAUs) without executing code. SAUs are organized into a four-category
diagnostic taxonomy covering numerical, methodological, protocol
and ordering drift. Repositories are scored against SAU claims using
dimension-specific LLM judges on a five-level rubric, producing a diagnostic
report with per-dimension scores rather than a single scalar.
Figure~\ref{fig:benchmark-overview} provides an overview.

We construct SA-Bench from 30 papers spanning ICLR, ICML and NeurIPS~2025 across five ML domains, yielding 1{,}491 human-verified SAU claims and evaluate 12 generator configurations (4 models $\times$ 3 scaffolds) for a total of 360 paper-level evaluations.

The main contributions of this work are as follows:
\begin{itemize}[leftmargin=1.5em, nosep]
  \item We formulate \emph{semantic alignment} as a measurable construct
    for paper-to-code reproduction, operationalized through Semantic
    Alignment Units (SAUs) and a four-category diagnostic taxonomy of
    semantic drift.
  \item We present \textsc{SemanticAlign-Bench}, a public benchmark of 30
    papers with 1{,}491 human-verified SAU claims, paired with an
    automated multi-agent scoring pipeline that enables scalable
    evaluation.
  \item We find that semantic drift is systematic across all tested
    configurations, and that scaffolds optimized for executability
    provide limited leverage for scientific reproduction, indicating
    that semantic specification verification is a critical direction
    for paper-to-code agents.
\end{itemize}

\section{Related Work}
\label{sec:related-work}

\noindent\textbf{Paper-to-code generation.}
Paper-to-code reproduction is a specialized subclass of
NL2Repo~\cite{nl2repobench}. The task asks an agent to read a research paper
and produce an executable repository that implements the paper.
PaperCoder~\cite{papercoder} decomposes the task into planning, analysis and
coding stages across three agent roles. RePro~\cite{repro} introduces atomic
verifiable criteria with reflective self-correction at each reproduction step.
AutoP2C~\cite{autop2c} structures reproduction as a
paper-understanding stage feeding code generation, and Lin et
al.~\cite{promptfree-collab} explore prompt-free collaborative multi-agent
reproduction. As generation methods improve, rigorous
evaluation of the resulting code becomes the bottleneck.

\noindent\textbf{Benchmarks for paper-to-code evaluation.}
Several benchmarks measure paper-to-code reproduction quality.
PaperBench~\cite{paperbench} decomposes 20 ICML~2024 papers into
author-co-developed rubric trees scored by an LLM judge against agent-generated
repositories. LMR-BENCH~\cite{lmrbench} and
SciReplicate-Bench~\cite{scireplicatebench} evaluate function-level
reproduction by executing unit tests on agent-implemented functions extracted
from research papers; SciCode~\cite{scicode} extends this to complex scientific
algorithms. PRBench~\cite{prbench} broadens reproduction evaluation to physics
research. CORE-Bench~\cite{corebench} measures computational reproducibility by
requiring agents to re-execute author-released code and recover reported
results. SciCoQA~\cite{scicoqa} formulates paper--code alignment as a
question-answering task. These benchmarks measure whether reproduction
succeeds and produce pass/fail or scalar scores at the document or function
level. None offer a systematic classification of the deviations between
generated code and paper specifications, leaving semantic drift undiagnosed
when reproduction fails.

SA-Bench addresses this gap by evaluating reproduction at the granularity of
individual implementation claims (Semantic Alignment Units), scoring each on a
five-level rubric and mapping deviations to a four-category drift taxonomy.
Table~\ref{tab:benchmark-comparison} positions SA-Bench against existing
benchmarks. This work connects to the broader literature on semantic drift in
long-form generation~\cite{know-when-to-stop}, goal drift and degradation in
coding agents~\cite{goal-drift,slopcodebench}, and LLM-based code
evaluation~\cite{codejudge,sedareval,llm-judge-survey}.
\section{SemanticAlign-Bench}
\label{sec:sa-bench}

We formalize semantic alignment evaluation as a static, claim-level scoring
problem and describe how \textsc{SemanticAlign-Bench} (SA-Bench) instantiates
it: we define the task (\S\ref{sec:task-formulation}), describe paper selection
(\S\ref{sec:paper-selection}), introduce the SAU framework and drift taxonomy
(\S\ref{sec:design}) and present the extraction and judging pipelines
(\S\ref{sec:sau-extraction}--\S\ref{sec:judge-scoring}).

\subsection{Task Formulation}
\label{sec:task-formulation}

The input is a research paper $P$ and an agent-generated code repository $R$
purporting to reproduce $P$. The output is a paper-level Semantic Alignment
Score with per-claim diagnostic judgments: for each
verifiable implementation claim extracted from $P$, we report whether $R$ faithfully
realizes it and, if not, what failure occurred. The scoring process is
\textbf{static}: it inspects paper text and code text without executing $R$,
isolating semantic fidelity from environmental factors.

\subsection{Paper Selection}
\label{sec:paper-selection}
SA-Bench comprises 30 papers from NeurIPS, ICML and ICLR~2025, spanning five domains and six research paradigms.
% The five domains are NLP/LLM, Computer Vision, Reinforcement Learning, Probabilistic Inference/Generative Models and Scientific Computing/Numerical Methods. The six paradigms are New Algorithm/Architecture, Incremental Improvement, Theoretical Analysis, System/Pipeline, Empirical Comparison and Generative Model.
The full paper list with venue, SAU counts, paradigm labels and per-dimension breakdown is
in Appendix~\ref{app:paper-list}.

\subsection{Design}
\label{sec:design}
\begin{figure}[h!]
\centering
\includegraphics[width=0.95\columnwidth]{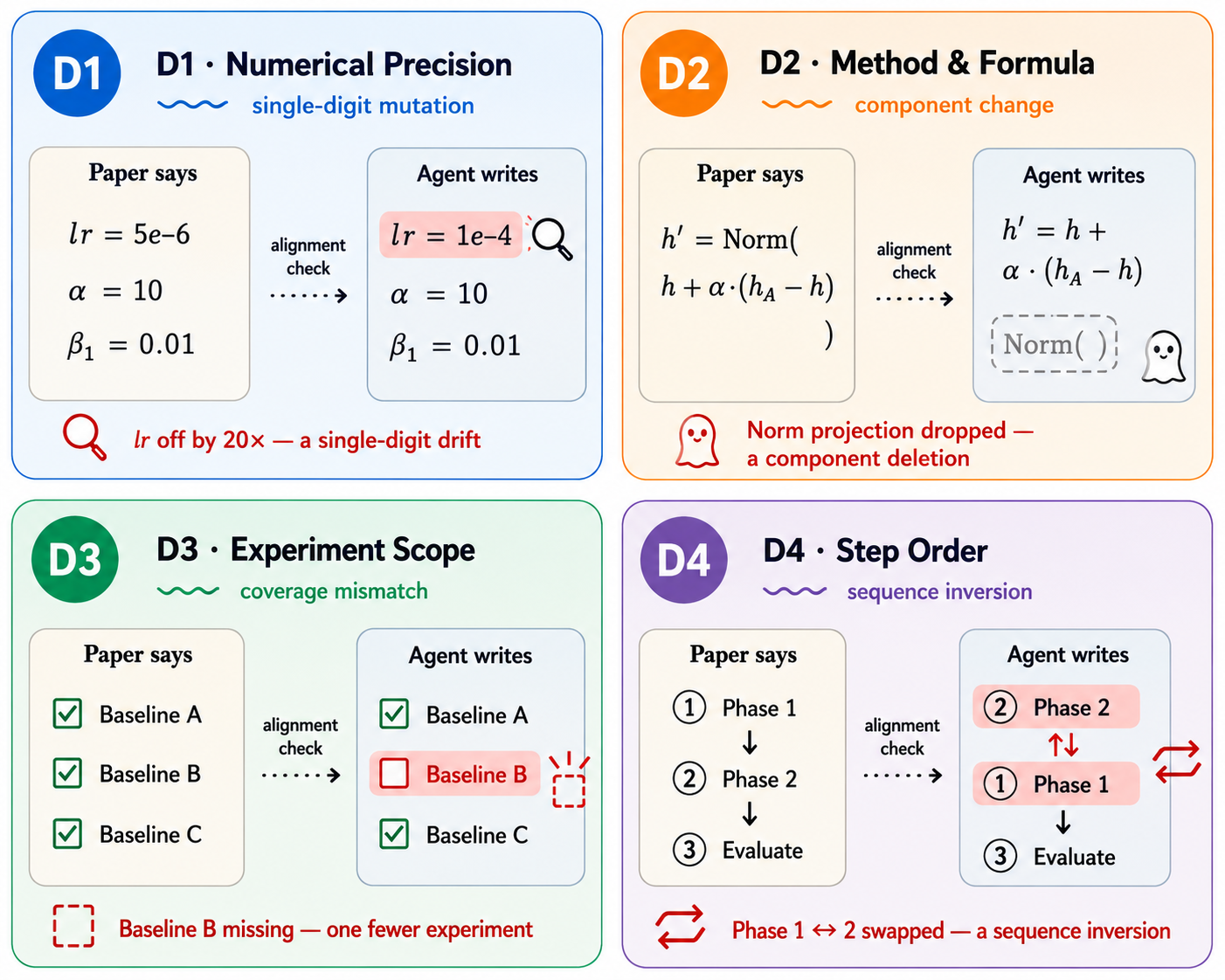}
\caption{Four diagnostic dimensions of semantic drift.}
\label{fig:drift-taxonomy}
\end{figure}

\begin{figure*}[t!]
\centering
\includegraphics[width=0.99\textwidth]{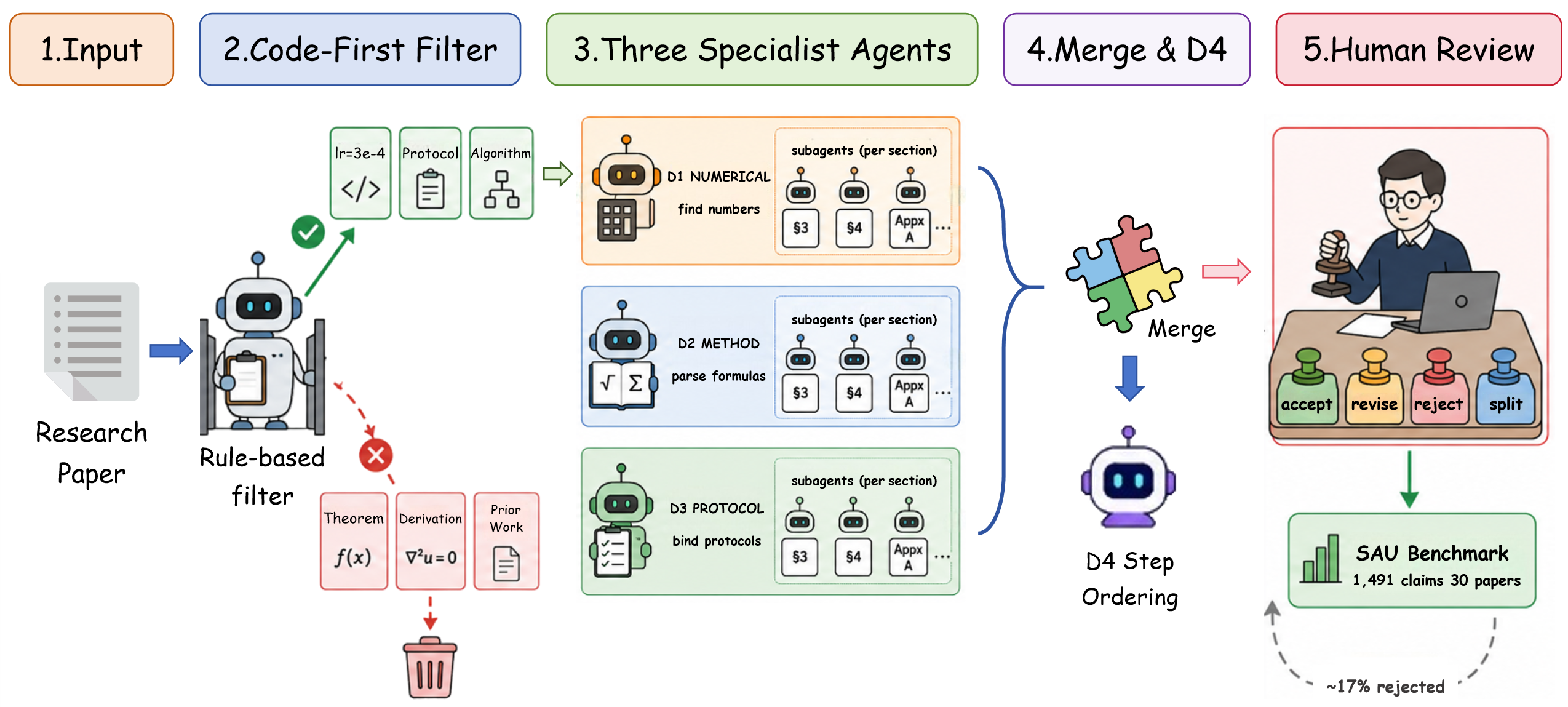}
\caption{SA-Bench construction pipeline. A code-first filter separates
implementable content from theoretical discussion; three specialized agents
(D1 Numerical, D2 Method, D3 Protocol) extract SAU claims in parallel with
per-section spawning to avoid attention collapse. D4 claims are derived from
D2/D3 ordering annotations during a merge phase. All claims undergo mandatory
human review before inclusion in the benchmark.}
\label{fig:sa-bench-pipeline}
\end{figure*}

\textbf{A Semantic Alignment Unit (SAU)} is
an atomic implementation claim extracted from $P$. Each SAU satisfies three
properties: (1)~\textbf{Localizable} to a concrete paper source span;
(2)~\textbf{Verifiable} by static code inspection;
(3)~\textbf{Atomic}, describing a single independently assessable implementation
decision. Full annotation schema and examples are in
Appendix~\ref{app:sau-example}.

\textbf{Drift taxonomy.} We categorize deviations from faithful implementation
into four diagnostic types, derived from a pilot study on
PaperBench-dev~\cite{paperbench}, detailed
in Appendix~\ref{app:pilot-study}.
Each type names the kind of paper specification the agent misreads, ordered from
value-level to procedural-level:

\begin{itemize}[leftmargin=1.5em, nosep]
  \item \textbf{D1 (Numerical Precision):} hyperparameters, thresholds, or
    architectural dimensions.
  \item \textbf{D2 (Method / Formula):} algorithm steps, equations, or
    component interfaces added, deleted, or replaced.
  \item \textbf{D3 (Experimental Protocol):} datasets, baselines, metrics,
    or ablations added or omitted.
  \item \textbf{D4 (Step Ordering):} phase sequences or pipeline steps
    transposed or merged, restricted to explicitly stated ordering
    constraints.
\end{itemize}

D1--D4 provide a systematic classification of the deviations themselves, as
Figure~\ref{fig:drift-taxonomy} illustrates.

\subsection{SAU Extraction}
\label{sec:sau-extraction}
Extracting SAU claims from a 20--40 page paper faces two obstacles: a single
LLM pass over the full paper suffers attention
decay~\cite{lost-in-the-middle,longbench} and paper text
entangles specifications, derivations and experimental results, while much of
this content does not require code implementation and is redundant for our
task. We operationalize extraction as a structured prompt-driven agent pipeline
followed by human review; Figure~\ref{fig:sa-bench-pipeline} illustrates the full
workflow.

\smallskip
\textbf{Pipeline.} As shown in Figure~\ref{fig:sa-bench-pipeline}, the pipeline has
four stages. A code-first filter first removes
non-implementable content without invoking an LLM. Three specialist agents then
process the paper in parallel under dimension-specific prompts, each
dispatching per-section sub-agents that read 1--3 sections at a time
to avoid long-context attention decay. A final merge stage
deduplicates outputs and derives D4 step-ordering claims from annotations
recorded by D2/D3, rather than from a separate extraction pass. Per-agent
prompts, decision trees and the merge protocol are in
Appendix~\ref{app:extraction-details}.

\begin{figure*}[t!]
\centering
\includegraphics[width=0.98\textwidth]{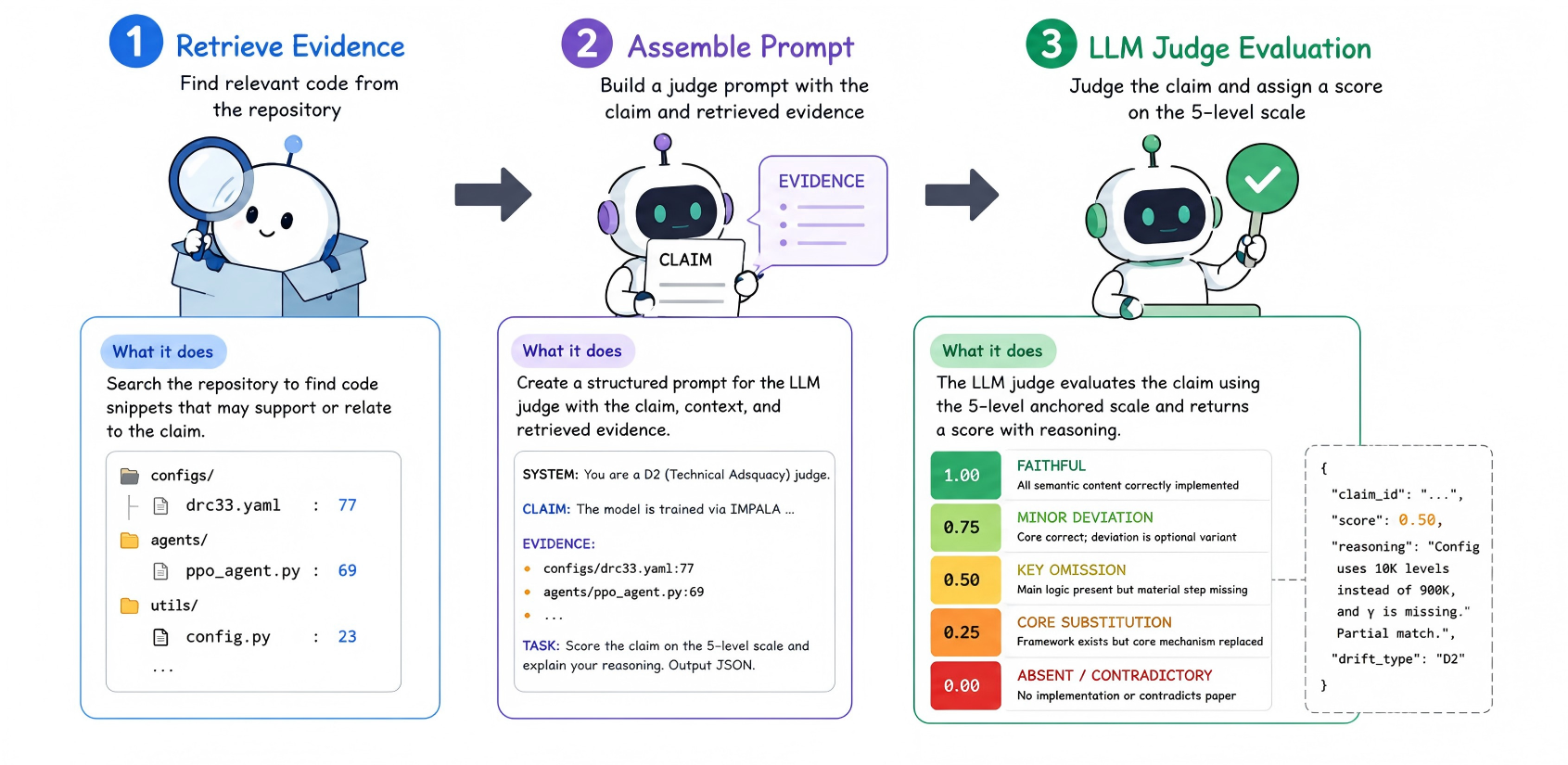}
\caption{Example SAU scoring output for a single claim (D1-006,
emergent-planning-rl). The judge outputs structured JSON with claim-level scores,
cited evidence with file:line references and per-dimension aggregates. Full
judge prompt templates are in Appendix~\ref{app:judge-prompts}.}
\label{fig:sau-score}
\end{figure*}
\smallskip
\textbf{Human review.} The pipeline favors recall over precision and relies on
human reviewers to filter and refine the extracted candidates. Reviewers check
each candidate against four criteria: contribution relevance, drift type
correctness, granularity and source precision.

They may accept, revise,
reject, or split a claim; every action is recorded with reasoning in a
per-paper review log. Across all 30 papers, approximately 17\% of
agent-extracted candidates are rejected and many more are revised. Review
outcomes include acceptance, revision or retyping, rejection or deletion,
merging or splitting, and addition of missed claims. The rejected and modified
candidates follow four recurring patterns: prior-work
contamination, where implementation details of cited methods are mistaken for
the target paper's contribution; result-value extraction, where reported
outcomes are treated as implementation requirements; D1--D4 type
confusion, where architecture or training decisions receive the wrong drift
label; and over-splitting, where coupled components are fragmented into
claims that are not independently implementable. The review log records these
actions with reviewer rationales; representative examples and per-paper
statistics are provided in Appendix~\ref{app:review-examples}. The final
benchmark contains 1{,}491 SAU claims across 30 papers.

\subsection{Judge Scoring}
\label{sec:judge-scoring}

\textbf{Scoring protocol.} We score each SAU on the five-level rubric in
Table~\ref{tab:scoring-scale}; detailed examples with per-type calibration
are in Appendix~\ref{app:scoring-rubric}.

\begin{table}[]
\centering
\footnotesize
\caption{Five-level scoring rubric.}
\label{tab:scoring-scale}
\begin{tabular}{@{}clp{3.8cm}@{}}
\toprule
\textbf{Score} & \textbf{Label} & \textbf{Criterion} \\
\midrule
1.0 & Faithful & All semantic content correctly implemented. \\
0.75 & Minor deviation & Core correct; deviation is an optimization variant or
  implementation detail. \\
0.5 & Key omission & Main logic present but a material step or component missing. \\
0.25 & Core substitution & Core mechanism replaced by a different approach. \\
0 & Absent & No implementation or logic directly contradicts paper. \\
\bottomrule
\end{tabular}
\end{table}

The paper-level SAS is the mean of individual SAU scores:
\begin{equation*}
\text{SAS}(P, R) =
\frac{1}{|\mathcal{S}(P)|} \sum_{s \in \mathcal{S}(P)} \text{score}(s, R)
\end{equation*}
where $\text{score}(s, R)$ $\in$ $\{0, 0.25, 0.5, 0.75, 1.0\}$ and $\mathcal{S}(P)$
includes all D1--D4 claims, with D4 scored using the same rubric as the other
dimensions. We additionally
report per-type averages $\text{SAS}_{\text{D1}}$, $\ldots$, $\text{SAS}_{\text{D4}}$
for diagnostic comparison.

\smallskip
\textbf{Pipeline.} Figure~\ref{fig:sau-score} illustrates the scoring workflow.
The judge (GPT-5.5) searches the repository for evidence matching each SAU claim,
filters out claims with no code-level support and scores the remainder using
dimension-specific prompts. Each scored claim is accompanied by a structured
explanation of the observed alignment or deviation.

\smallskip
\textbf{Judge correctness.} To assess scoring reliability, we randomly sampled
200 claims across all dimensions and generators and manually verified each
score against the evidence. The judge's score matched the human assessment
in $\sim$87\% of cases. Disagreements concentrated
in D2 claims where the boundary between ``core mechanism present but
incomplete'' (0.5) and ``core mechanism missing'' (0.25) is ambiguous. Full
rubric tables and scoring examples are in
Appendix~\ref{app:scoring-rubric}; the complete judge output schema is in
Appendix~\ref{app:judge-output-schema}.

\section{Evaluation Setup}
\label{sec:evaluation-setup}

We evaluate 12 generator configurations (4 models $\times$ 3 scaffolds) over
all 30 papers, yielding 360 paper-level evaluations and enabling joint
analysis of model capability and scaffold strategy on semantic alignment.

\textbf{Models.}  We use Claude-Sonnet-4.6~\cite{claude-sonnet-4-6}\footnote{\url{https://www.anthropic.com/claude/sonnet}},
  DeepSeek-V4-Pro~\cite{deepseek-v4}\footnote{\url{https://api-docs.deepseek.com/news/news260424}},
  Gemini-2.5-Flash~\cite{gemini-2-5}\footnote{\url{https://ai.google.dev/gemini-api/docs/models/gemini-2.5-flash-preview-09-2025}}
  and GPT-4o~\cite{gpt-4o}\footnote{\url{https://platform.openai.com/docs/models/gpt-4o}}, covering
  four model families spanning multiple capability
  tiers; all share the same tool interface and identical paper
inputs.

\textbf{Scaffolds.} We compare three scaffolds representing distinct
agent paradigms: a minimal ReAct baseline, a dedicated paper-to-code
pipeline and a software-engineering execution scaffold.

\emph{BasicAgent}, following the design in \citet{paperbench}, is a
minimal ReAct loop with shell, local file read and
local-file regex search tools.

\emph{PaperCoder}~\cite{papercoder} is a dedicated paper2code framework
with a three role pipeline consisting of planning, analysis and coding.

\emph{OpenHands}~\cite{openhands} is a widely-used software-engineering
scaffold with a CodeAct execution-feedback loop consisting of action,
execution and observation over shell and file-editor tools.
Per-scaffold hyperparameters and tool inventories are listed in
Appendix~\ref{app:scaffold-config}.

\textbf{Protocol.} Each (paper, repository) pair is scored by the SAU
scoring pipeline of Section~\ref{sec:judge-scoring} with GPT-5.5~\cite{gpt-5-5}\footnote{\url{https://openai.com/index/introducing-gpt-5-5/}} as judge
to prevent self-evaluation bias. The primary metric is paper-level SAS;
diagnostic metrics include per-type sub-scores
$\text{SAS}_{\text{D1}}$--$\text{SAS}_{\text{D4}}$ and a zero-score
decomposition.

\section{Results and Analysis}
\label{sec:results}

We report results from 360 paper-level evaluations across 12 generators and 30
papers, totaling 17,892 claim-level judgments.

\subsection{Overall Performance}
\label{sec:results-overall}

Performance is uniformly low: the overall SAS mean across all 360 evaluations
is 0.221, with median 0.237. The best single configuration,
Claude-Sonnet-4.6 + PaperCoder, reaches 0.301.
Figure~\ref{fig:generator-ranking} shows the per-paper score distribution.
Full per-configuration scores are in Appendix~\ref{app:full-results}.

\begin{figure}[t]
\centering
\includegraphics[width=0.88\columnwidth]{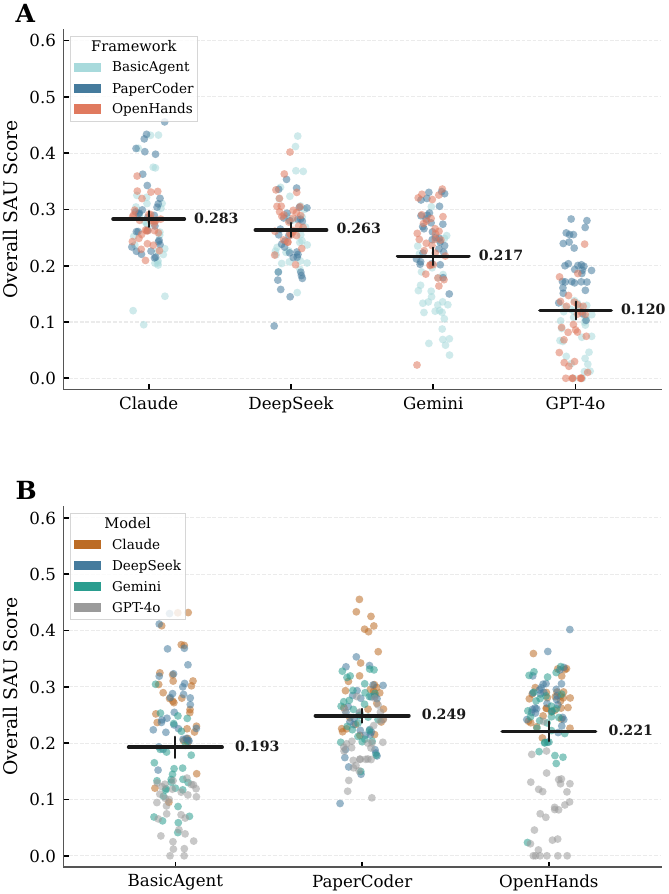}
\caption{Overall SAS distribution by model and scaffold. Each dot is one paper;
crossbars show mean $\pm$ 95\% CI.}
\label{fig:generator-ranking}
\end{figure}

\textbf{Model choice outweighs scaffold choice on average, but the pattern is not uniform.}
The top five configurations all use Claude or DeepSeek; GPT-4o + PaperCoder
scores 0.197, below every Claude/DeepSeek configuration. Scaffold benefit
varies with base capability, as Table~\ref{tab:scaffold-delta} shows: PaperCoder helps
weaker models most and saturates for stronger ones, with DeepSeek a slight
regression. OpenHands shows a notable gain only on Gemini ($+$0.093) with minimal effect
elsewhere ($\leq$$+$0.014). For PaperCoder, the trend is clearer: benefit decreases as base capability
increases. GPT-4o and Gemini have low bare baselines; without scaffolding,
they struggle to structure the task. PaperCoder's analyze-then-code template
compensates for this gap, yielding large gains. Claude and DeepSeek already
score well without scaffolding: they natively extract specifications, plan
implementations and write code aligned to paper content. A fixed pipeline instead crowds out their own
effective strategies.

\begin{table}[t]
\centering
\scriptsize
\setlength{\tabcolsep}{4pt}
\caption{Mean SAS per model and scaffold. Deltas relative to BasicAgent:
\textcolor{red}{red} $=$ gain, \textcolor{green!60!blue}{green} $=$ regression.}
\label{tab:scaffold-delta}
\begin{tabular}{@{}lrrr@{}}
\toprule
\textbf{Model} & \textbf{BasicAgent} & \textbf{PaperCoder} & \textbf{OpenHands} \\
\midrule
Claude-Sonnet-4.6   & 0.272 & 0.301\;\textcolor{red}{\scriptsize (+0.029)} & 0.277\;\textcolor{red}{\scriptsize (+0.005)} \\
DeepSeek-V4-Pro & 0.268 & 0.241\;\textcolor{green!60!blue}{\scriptsize ($-$0.027)} & 0.282\;\textcolor{red}{\scriptsize (+0.014)} \\
Gemini-2.5-Flash   & 0.150 & 0.256\;\textcolor{red}{\scriptsize (+0.106)} & 0.243\;\textcolor{red}{\scriptsize (+0.093)} \\
GPT-4o   & 0.081 & 0.197\;\textcolor{red}{\scriptsize (+0.116)} & 0.082\;\textcolor{red}{\scriptsize (+0.001)} \\
\bottomrule
\end{tabular}
\end{table}

\begin{table*}[t]
\centering
\footnotesize
\caption{Marginal means by model and scaffold across all four drift dimensions.
Model range: 0.120--0.283; scaffold range: 0.193--0.249.}
\label{tab:model-scaffold-marginals}
\begin{tabular}{@{}lrrrrr@{}}
\toprule
\textbf{Factor} & \textbf{Overall} & \textbf{D1} & \textbf{D2} & \textbf{D3} & \textbf{D4} \\
\midrule
\multicolumn{6}{c}{\textit{By Model}} \\
Claude-Sonnet-4.6   & 0.283 & 0.413 & 0.293 & 0.201 & 0.225 \\
DeepSeek-V4-Pro     & 0.263 & 0.358 & 0.294 & 0.186 & 0.216 \\
Gemini-2.5-Flash    & 0.217 & 0.268 & 0.248 & 0.161 & 0.189 \\
GPT-4o              & 0.120 & 0.120 & 0.140 & 0.092 & 0.130 \\
\midrule
\multicolumn{6}{c}{\textit{By Scaffold}} \\
PaperCoder          & 0.249 & 0.344 & 0.262 & 0.179 & 0.211 \\
OpenHands           & 0.221 & 0.288 & 0.233 & 0.174 & 0.188 \\
BasicAgent          & 0.193 & 0.238 & 0.237 & 0.127 & 0.170 \\
\bottomrule
\end{tabular}
\end{table*}

\subsection{Dimension Hierarchy}
\label{sec:results-dimension}

\begin{figure}[h]
\centering
\includegraphics[width=0.88\columnwidth]{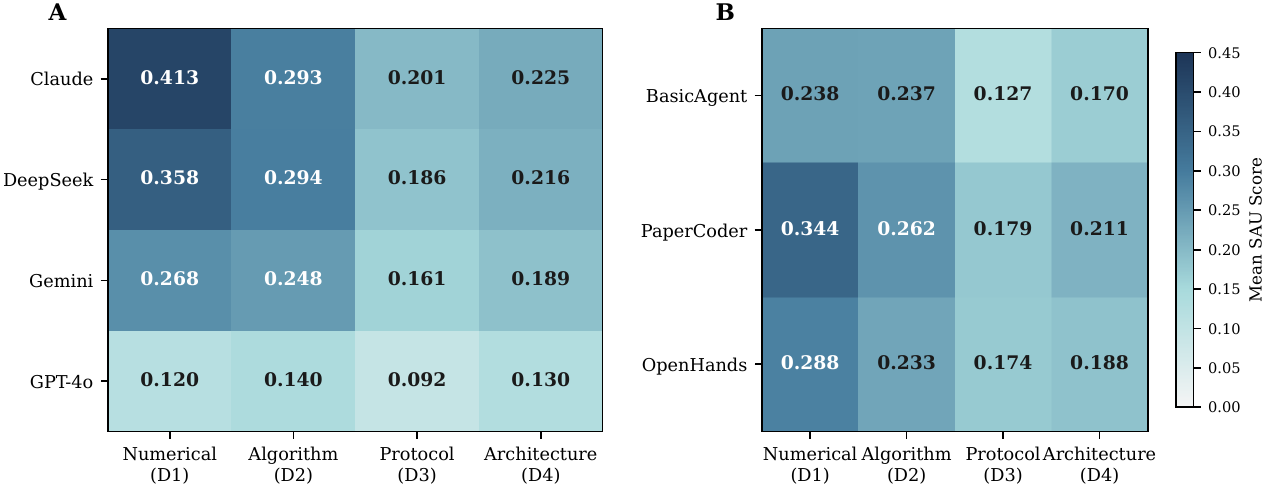}
\caption{Per-dimension mean SAU score heatmaps}
\label{fig:dimension-profile}
\end{figure}

\begin{table*}[t]
\centering
\footnotesize
\caption{Mean SAS for all 12 generator configurations, grouped by scaffold.
\textbf{Bold} marks the best score per column; \underline{underline} marks the
second best.}
\label{tab:generator-scores-full}
\begin{tabular}{@{}llrrrrr@{}}
\toprule
\textbf{Scaffold} & \textbf{Model} & \textbf{Overall} & \textbf{D1} & \textbf{D2} & \textbf{D3} & \textbf{D4} \\
\midrule
            & Claude-Sonnet-4.6  & 0.272 & \underline{0.391} & \underline{0.307} & 0.169 & 0.219 \\
BasicAgent  & DeepSeek-V4-Pro    & 0.268 & 0.379 & \textbf{0.308} & 0.180 & 0.204 \\
            & Gemini-2.5-Flash   & 0.150 & 0.134 & 0.210 & 0.099 & 0.158 \\
            & GPT-4o             & 0.081 & 0.047 & 0.121 & 0.058 & 0.100 \\
\midrule
            & Claude-Sonnet-4.6  & \textbf{0.301} & \textbf{0.470} & 0.299 & 0.209 & 0.226 \\
PaperCoder  & DeepSeek-V4-Pro    & 0.241 & 0.339 & 0.278 & 0.148 & 0.200 \\
            & Gemini-2.5-Flash   & 0.256 & 0.335 & 0.270 & 0.198 & 0.222 \\
            & GPT-4o             & 0.198 & 0.231 & 0.200 & 0.163 & 0.197 \\
\midrule
            & Claude-Sonnet-4.6  & 0.276 & 0.379 & 0.272 & \underline{0.224} & \underline{0.229} \\
OpenHands   & DeepSeek-V4-Pro    & \underline{0.282} & 0.357 & 0.297 & \textbf{0.229} & \textbf{0.244} \\
            & Gemini-2.5-Flash   & 0.243 & 0.335 & 0.264 & 0.187 & 0.187 \\
            & GPT-4o             & 0.082 & 0.082 & 0.098 & 0.056 & 0.093 \\
\midrule
\multicolumn{2}{l}{\textit{Mean}} & 0.221 & 0.290 & 0.244 & 0.160 & 0.190 \\
\bottomrule
\end{tabular}
\end{table*}

The D1 $>$ D2 $>$ D4 $>$ D3 ordering holds across all 12 configurations
and is universal across models and scaffolds (full statistics in
Table~\ref{tab:dimension-stats-full}). The hierarchy tracks the size of
what each claim demands: D1 asks for a single value, trivially copyable;
D2 asks for a formula or algorithm step, implementable within a few lines;
D4 asks for a pipeline ordering, satisfiable as long as the sequence is
right; D3 asks for a full component (a baseline, a dataset loader, an
evaluation protocol), each comprising multiple sub-decisions that must all
be correct. Each additional D3 constraint compounds the burden: with more
pieces that can go wrong, the probability of a perfect implementation
drops, producing the universal bottleneck at D3.

\subsection{Domain and Paradigm Analysis}
\label{sec:results-domain}

\begin{figure}[h]
\centering
\includegraphics[width=0.98\columnwidth]{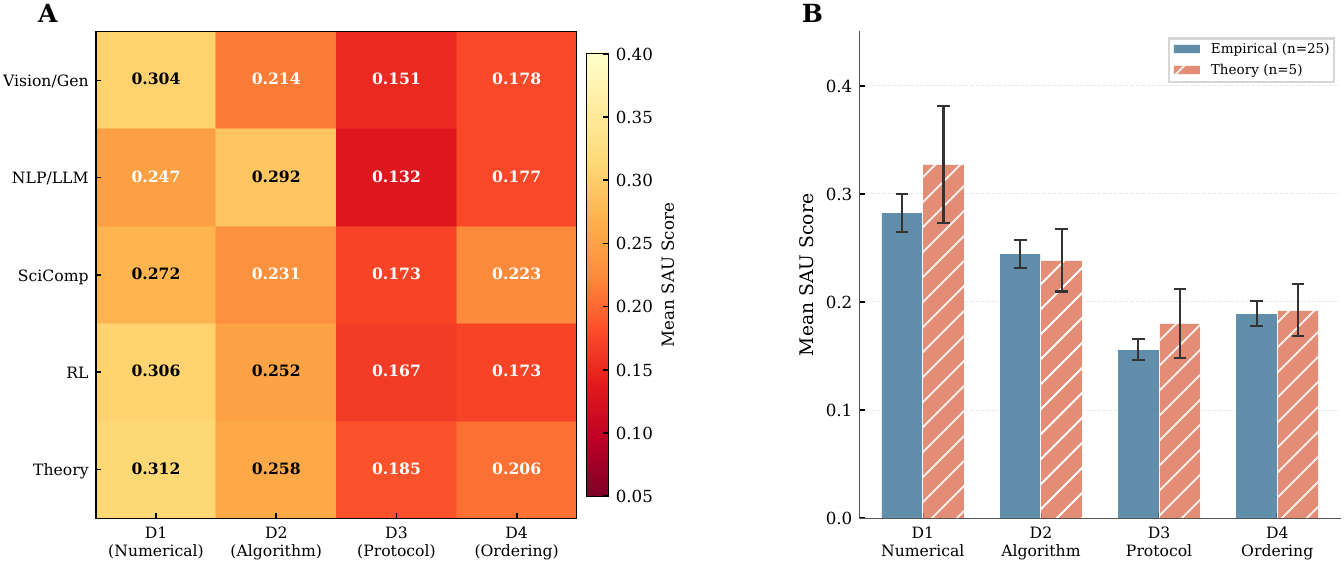}
\caption{Domain $\times$ Dimension heatmap of mean SAU scores across all
12 generator configurations.}
\label{fig:domain-paradigm}
\end{figure}

Figure~\ref{fig:domain-paradigm} shows the domain $\times$ dimension
breakdown. The cross-domain spread in overall SAS is 0.030, modest relative
to the 0.219 gap between best and worst configurations: generator quality
dominates domain effects.

Domain differences are dimension-specific. NLP/LLM leads on D2, consistent
with explicit pseudocode and equation blocks common in language modeling.
Probabilistic Inference/Generative Models leads on D1, reflecting detailed
hyperparameter tables for sampling and training. Computer Vision trails on
D3 and D4, where multi-stage training protocols are hard to recover from
prose.

Across paradigms, the spread is similarly narrow. One asymmetry worth
noting: Empirical Comparison papers attain strong D2 scores because they
re-cite existing methods with explicit formula blocks, yet their D3 scores
are among the lowest as protocol details are spread across appendices.
Full per-paradigm scores are in Table~\ref{app:paradigm-scores}.

\subsection{Where Zero Scores Come From}
\label{sec:results-error}

\begin{figure}[h]
\centering
\includegraphics[width=0.85\columnwidth]{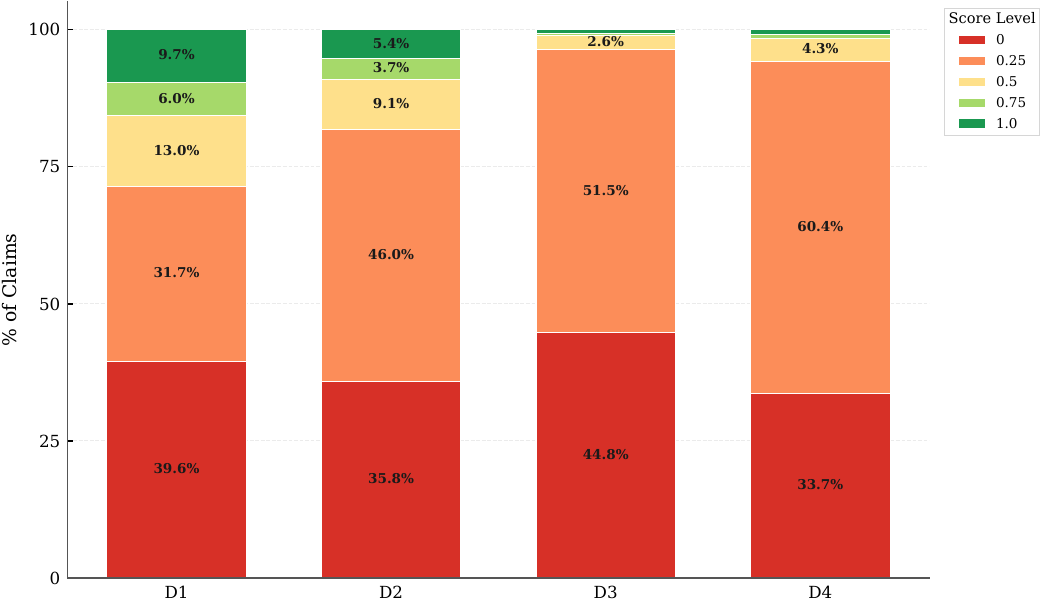}
\caption{Score-level distribution of all 17,892 claim judgments}
\label{fig:error-breakdown}
\end{figure}

Figure~\ref{fig:error-breakdown} shows the full score distribution.
To examine where zero scores come from, we classify all 7,034 zero-scored
claims by keyword-matching the judge's natural-language reasoning against a
small lexicon defined in Appendix~\ref{app:case-analysis}.

Three categories cover the bulk of zero-scored claims:

\begin{enumerate}[leftmargin=1.5em, nosep]
  \item \textbf{Implementation mismatch, 40.8\%.} The judge located code
    referencing the claim's keywords, but determined the implementation does
    not match the specified content: the agent wrote something unrelated
    under recognizable names.
  \item \textbf{Stub or placeholder, 16.2\%.} Evidence exists but is a stub,
    a TODO, or a \texttt{pass} statement: the agent acknowledged the requirement
    but deferred implementation.
  \item \textbf{External knowledge gap, 8.0\%.} The claim requires
    implementing a standard baseline like PPO or loading a benchmark dataset
    like MOSE, which the paper names but does not define.
\end{enumerate}

A finer-grained six-type taxonomy with examples is provided in
Appendix~\ref{app:case-analysis}, covering token mention only, wrong
implementation, missing component and unclassified cases.

\subsection{Discussion: Why Scaffolds Underperform and What to Build Next}
\label{sec:results-discussion}

\textbf{Existing scaffolds optimize for the wrong signal.}
OpenHands uses CodeAct: the agent writes code, executes it, observes test
failures and stack traces and iterates. This loop is coupled to
\emph{executability}. On SWE-Bench~\cite{swebench}, where the task is
fixing bugs in an existing codebase, execution feedback is directly aligned
with the goal. On SA-Bench, code can execute cleanly while implementing the
wrong algorithm.

PaperCoder separates paper analysis from code generation, which helps
ground code in paper specifications. However, it remains a one-pass
pipeline: the reviewer checks code quality but does not verify that the
generated code faithfully implements the extracted specifications. Without
a semantic verification loop, its gains are inherently bounded. Together,
these observations suggest that semantic specification verification is a
key direction for narrowing the semantic-alignment gap in future
paper-to-code agents.

\FloatBarrier

\section{Conclusion}

We introduce \textsc{SemanticAlign-Bench}, a diagnostic benchmark that evaluates
whether agent-generated code faithfully implements a paper's specifications,
with a human-verified multi-agent pipeline producing 1{,}491 Semantic Alignment
Units (SAUs) across 30 papers from five ML domains. Across 360 evaluations
spanning four models and three scaffolds, semantic drift is systematic (mean SAS
0.221) and the best configuration reaches only 0.301. A failure taxonomy reveals
that agents attempt most requirements but implement them incorrectly, with
implementation mismatch and stubs accounting for the majority of zero-scored
claims. Our analysis indicates that, for capable base models, scaffolds
designed for general software engineering provide limited leverage for
scientific reproduction, where executability alone is insufficient; closing
the gap requires scaffolds that prioritize semantic specification verification.

\section{Limitations}
\label{sec:limitations}

\textbf{Annotation cost.} Ensuring SAU claim fidelity requires expert review
against paper sources, which limits scalability. While the multi-agent
extraction pipeline reduces manual effort, developing automatic or
semi-automatic verification mechanisms would enable broader benchmark coverage.

\noindent\textbf{Domain scope.} SA-Bench focuses on 30 papers from ICLR, ICML and
NeurIPS 2025 across five ML domains. Findings may not directly generalize to
other fields such as biomedicine or theoretical disciplines.
Extending to these settings requires new paper curation and domain-specific
adaptation of the extraction pipeline, as code implementation patterns differ
across fields.

%\section*{Acknowledgments}
%This work was supported by the Fundamental and Interdisciplinary Disciplines Breakthrough Plan of the Ministry of Education of China (JYB2025XDXM113), the National Natural Science Foundation of China (92470121, 62402016), the National Key R\&D Program of China (2024YFA1014003), Zhongguancun Academy (C20250204, C20250602), the Beijing Major Science and Technology Project (Z251100008125043, Z251100008425023), and the High-performance Computing Platform of Peking University.

\newpage
\makeatletter
\let\hyper@@link\SA@hyper@@link
\let\hyper@link@\SA@hyper@link@
\let\hyper@link\SA@hyper@link
\let\hyper@linkurl\SA@hyper@linkurl
\let\hyper@linkstart\SA@hyper@linkstart
\let\hyper@linkend\SA@hyper@linkend
\makeatother
\bibliography{refs}

\clearpage
\appendix
\renewcommand{\appendixspacing}{%
  \setlength{\intextsep}{4pt plus 1pt minus 1pt}%
  \setlength{\textfloatsep}{4pt plus 1pt minus 1pt}%
  \setlength{\floatsep}{4pt plus 1pt minus 1pt}%
  \setlength{\dbltextfloatsep}{4pt plus 1pt minus 1pt}%
  \setlength{\dblfloatsep}{4pt plus 1pt minus 1pt}%
  \setlength{\abovecaptionskip}{2pt}%
  \setlength{\belowcaptionskip}{0pt}%
}
\appendixspacing
\setlength{\LTpre}{0pt}
\setlength{\LTpost}{0pt}
\setcounter{topnumber}{5}
\setcounter{dbltopnumber}{5}
\setcounter{totalnumber}{8}
\renewcommand{\topfraction}{0.95}
\renewcommand{\dbltopfraction}{0.95}
\renewcommand{\textfraction}{0.05}
\renewcommand{\floatpagefraction}{0.75}
\renewcommand{\dblfloatpagefraction}{0.75}

% EMNLP/ACL appendices remain in two-column format.  Longtable cannot run in
% two-column mode, so only the oversized appendix tables are isolated on
% one-column pages and normal appendix text/tables resume two-column layout.
\makeatletter
\setlength{\@fptop}{0pt}
\setlength{\@fpsep}{6pt plus 2pt minus 2pt}
\setlength{\@fpbot}{0pt plus 1fil}
\setlength{\@dblfptop}{0pt}
\setlength{\@dblfpsep}{6pt plus 2pt minus 2pt}
\setlength{\@dblfpbot}{0pt plus 1fil}
\def\section{\@startsection {section}{1}{\z@}{-1.4ex plus
    -0.3ex minus -.1ex}{0.8ex plus 0.15ex minus .1ex}{\large\bfseries\raggedright}}
\def\subsection{\@startsection{subsection}{2}{\z@}{-1.1ex plus
    -0.3ex minus -.1ex}{0.35ex plus .1ex minus .1ex}{\normalsize\bfseries\raggedright}}
\def\subsubsection{\@startsection{subsubsection}{3}{\z@}{-0.9ex plus
   -0.2ex minus -.1ex}{0.25ex plus .1ex minus .1ex}{\normalsize\bfseries\raggedright}}
\def\paragraph{\@startsection{paragraph}{4}{\z@}{0.7ex plus
   0.2ex minus .1ex}{-1em}{\normalsize\bfseries}}
\newif\ifappendixreturntotwocolumn
\appendixreturntotwocolumntrue
\let\appendix@longtable\longtable
\let\endappendix@longtable\endlongtable
\renewenvironment{longtable}[1]{%
  \if@twocolumn
    \appendixreturntotwocolumntrue
    \onecolumn
    \appendixspacing
  \else
    \appendixreturntotwocolumnfalse
    \appendixspacing
  \fi
  \appendix@longtable{#1}%
}{%
  \endappendix@longtable
  \ifappendixreturntotwocolumn
    \twocolumn
    \appendixspacing
  \fi
}
\makeatother

% Prompt-style examples are inline, breakable blocks, not floats.  They remain
% in the current two-column flow and temporarily suspend review line numbering
% so long code/prompt examples do not consume reviewer line numbers.
\definecolor{reproboxbg}{RGB}{248,250,252}
\definecolor{reproboxsoft}{RGB}{245,249,255}
\definecolor{reproboxframe}{RGB}{114,139,176}
\definecolor{repropromptbg}{RGB}{246,250,255}
\definecolor{repropromptframe}{RGB}{91,126,173}
\renewenvironment{promptbox}[2]{%
  \par\smallskip
  \nolinenumbers
  \begin{tcolorbox}[
    enhanced jigsaw,
    breakable,
    colback=repropromptbg,
    colframe=repropromptframe,
    colbacktitle=repropromptbg,
    coltitle=black,
    fonttitle=\small\bfseries,
    title={#1},
    label={#2},
    boxrule=0.6pt,
    titlerule=0.3pt,
    arc=1.5mm,
    left=1mm,
    right=1mm,
    top=1mm,
    bottom=1mm,
    boxsep=1mm,
    before upper={\ttfamily\footnotesize\raggedright\sloppy
      \setlength{\parindent}{0pt}\setlength{\parskip}{0pt}},
    before skip=0pt,
    after skip=0pt
  ]%
}{%
  \end{tcolorbox}%
  \linenumbers
  \smallskip
}
\renewenvironment{jsonbox}[2]{%
  \par\smallskip
  \nolinenumbers
  \begin{tcolorbox}[
    enhanced jigsaw,
    breakable,
    colback=reproboxsoft,
    colframe=reproboxframe,
    colbacktitle=reproboxsoft,
    coltitle=black,
    fonttitle=\small\bfseries,
    title={#1},
    label={#2},
    boxrule=0.6pt,
    titlerule=0.3pt,
    arc=1.5mm,
    left=1mm,
    right=1mm,
    top=1mm,
    bottom=1mm,
    boxsep=1mm,
    before upper={\ttfamily\footnotesize\raggedright\sloppy
      \setlength{\parindent}{0pt}\setlength{\parskip}{0pt}},
    before skip=0pt,
    after skip=0pt
  ]%
}{%
  \end{tcolorbox}%
  \linenumbers
  \smallskip
}

% Listing 8 is kept intact on one page so its schema is not split across
% columns; other JSON examples retain the breakable jsonbox behavior.
\newenvironment{singlejsonbox}[2]{%
  \par\smallskip
  \nolinenumbers
  \begin{tcolorbox}[
    enhanced jigsaw,
    colback=reproboxsoft,
    colframe=reproboxframe,
    colbacktitle=reproboxsoft,
    coltitle=black,
    fonttitle=\small\bfseries,
    title={#1},
    label={#2},
    boxrule=0.6pt,
    titlerule=0.3pt,
    arc=1.5mm,
    left=1mm,
    right=1mm,
    top=1mm,
    bottom=1mm,
    boxsep=1mm,
    before upper={\ttfamily\fontsize{7pt}{7.2pt}\selectfont\raggedright\sloppy
      \setlength{\parindent}{0pt}\setlength{\parskip}{0pt}},
    before skip=0pt,
    after skip=0pt
  ]%
}{%
  \end{tcolorbox}%
  \linenumbers
  \smallskip
}
\renewenvironment{reviewbox}[2]{%
  \par\smallskip
  \nolinenumbers
  \begin{tcolorbox}[
    enhanced jigsaw,
    breakable,
    colback=reproboxbg,
    colframe=reproboxframe,
    colbacktitle=reproboxbg,
    coltitle=black,
    fonttitle=\small\bfseries,
    title={#1},
    label={#2},
    boxrule=0.6pt,
    titlerule=0.3pt,
    arc=1.5mm,
    left=1mm,
    right=1mm,
    top=1mm,
    bottom=1mm,
    boxsep=1mm,
    before upper={\small\raggedright\sloppy
      \setlength{\parindent}{0pt}\setlength{\parskip}{2pt}},
    before skip=0pt,
    after skip=0pt
  ]%
}{%
  \end{tcolorbox}%
  \linenumbers
  \smallskip
}

% ===================================================================
%  APPENDIX A: METHODS — Extraction pipeline and judge prompts
% ===================================================================

\onecolumn
\appendixspacing
\section{Benchmark Paper List}
\label{app:paper-list}

Table~\ref{tab:paper-list} lists all 30 papers in \textsc{SemanticAlign-Bench}
with their venue, total SAU count and per-type breakdown.

\begingroup
\small
\sloppy
\setlength{\LTpre}{3pt}
\setlength{\LTpost}{0pt}
\setlength{\tabcolsep}{1.6pt}
\renewcommand{\arraystretch}{0.90}
\begin{longtable}{@{}p{2.25cm}p{7.35cm}lccccc@{}}
\caption{Full benchmark paper list with SAU counts by drift type.}
\label{tab:paper-list}
\endfirsthead

\caption[]{Full benchmark paper list with SAU counts by drift type (continued).} \\
\toprule
\textbf{Paper ID} & \textbf{Title} & \textbf{Venue} & \textbf{Total} & \textbf{D1} & \textbf{D2} & \textbf{D3} & \textbf{D4} \\
\midrule
\endhead

\bottomrule
\multicolumn{8}{r}{\scriptsize\textit{Continued on next page}} \\
\endfoot

\bottomrule
\endlastfoot

\toprule
\textbf{Paper ID} & \textbf{Title} & \textbf{Venue} & \textbf{Total} & \textbf{D1} & \textbf{D2} & \textbf{D3} & \textbf{D4} \\
\midrule
adjoint-matching  & Adjoint Matching: Fine-tuning Flow and Diffusion Generative Models with Memoryless Stochastic Optimal Control  & ICLR 2025  & 47  & 17 & 20 & 7 & 3 \\
avg-reward-pg  & Global Convergence of Policy Gradient in Average Reward MDPs  & ICLR 2025  & 22  & 9 & 7 & 3 & 3 \\
ca2-vdm  & Ca2-VDM: Efficient Autoregressive Video Diffusion Model with Causal Generation and Cache Sharing  & ICML 2025  & 48  & 22 & 10 & 8 & 8 \\
cara  & Canonical Rank Adaptation: An Efficient Fine-Tuning Strategy for Vision Transformers  & ICML 2025  & 41  & 19 & 11 & 10 & 1 \\
conformal-bayesian-quadrature  & Conformal Prediction as Bayesian Quadrature  & ICML 2025  & 38  & 25 & 6 & 3 & 4 \\
diffusion-convergence-rate  & Instance-dependent Convergence Theory for Diffusion Models  & ICLR 2025  & 42  & 22 & 12 & 4 & 4 \\
emergent-planning-rl  & Interpreting Emergent Planning in Model-Free Reinforcement Learning  & ICLR 2025  & 60  & 20 & 17 & 17 & 6 \\
gated-attention-llm  & Gated Attention for Large Language Models  & NeurIPS 2025  & 61  & 15 & 19 & 17 & 10 \\
generator-augmented-flows  & Improving Consistency Models with Generator-Augmented Flows  & ICML 2025  & 54  & 14 & 26 & 12 & 2 \\
hi-mar  & Hierarchical Masked Autoregressive Models with Low-Resolution Token Pivots  & ICML 2025  & 50  & 28 & 12 & 6 & 4 \\
lora-sb  & Initialization Using Update Approximation is a Silver Bullet for Extremely Efficient Low-Rank Adaptation  & ICLR 2025  & 37  & 12 & 14 & 8 & 3 \\
luno  & Linearization Turns Neural Operators into Function-Valued Gaussian Processes  & ICML 2025  & 46  & 10 & 24 & 6 & 6 \\
ma-rlhf  & MA-RLHF: Reinforcement Learning from Human Feedback with Macro Actions  & ICLR 2025  & 44  & 9 & 19 & 11 & 5 \\
masked-diffusion-token-ordering  & Train for the Worst, Plan for the Best: Understanding Token Ordering in Masked Diffusion Models  & ICML 2025  & 65  & 28 & 24 & 9 & 4 \\
moe-pot  & Mixture-of-Experts Operator Transformer for Large-Scale PDE Pre-Training  & NeurIPS 2025  & 61  & 22 & 15 & 17 & 7 \\
mrq  & Towards General-Purpose Model-Free Reinforcement Learning  & ICLR 2025  & 35  & 9 & 17 & 5 & 4 \\
navil  & NaViL: Rethinking Scaling Properties of Native Multimodal Large Language Models  & NeurIPS 2025  & 43  & 11 & 12 & 10 & 10 \\
neural-operator-flow-matching-pde  & Bridging Neural Operator and Flow Matching for a Generative PDE Foundation Model  & NeurIPS 2025  & 38  & 13 & 15 & 6 & 4 \\
nfig  & NFIG: Multi-Scale Autoregressive Image Generation via Frequency Ordering  & NeurIPS 2025  & 30  & 7 & 9 & 7 & 7 \\
ngpt  & nGPT: Normalized Transformer with Representation Learning on the Hypersphere  & ICLR 2025  & 60  & 21 & 25 & 7 & 7 \\
olmoe  & OLMoE: Open Mixture-of-Experts Language Models  & ICLR 2025  & 58  & 20 & 29 & 6 & 3 \\
prioritized-generative-replay  & Prioritized Generative Replay  & ICLR 2025  & 57  & 23 & 14 & 17 & 3 \\
pyramidal-flow-matching  & Pyramidal Flow Matching for Efficient Video Generative Modeling  & ICLR 2025  & 69  & 32 & 17 & 9 & 11 \\
robotic-world-model  & Robotic World Model: A Neural Network Simulator for Robust Policy Optimization  & NeurIPS 2025  & 54  & 20 & 21 & 10 & 3 \\
sam2  & SAM 2: Segment Anything in Images and Videos  & ICLR 2025  & 77  & 28 & 23 & 21 & 5 \\
sc-fno  & Sensitivity-Constrained Fourier Neural Operators for Forward and Inverse Problems  & ICLR 2025  & 42  & 12 & 13 & 14 & 3 \\
score  & Training Language Models to Self-Correct via Reinforcement Learning  & ICLR 2025  & 40  & 6 & 13 & 9 & 12 \\
universal-neural-operators  & Towards Universal Neural Operators through Multiphysics Pretraining  & NeurIPS 2025  & 34  & 6 & 17 & 6 & 5 \\
voting-leaderboards  & Exploring and Mitigating Adversarial Manipulation of Voting-Based Leaderboards  & ICML 2025  & 44  & 13 & 10 & 9 & 12 \\
wdno  & Wavelet Diffusion Neural Operator  & ICLR 2025  & 94  & 30 & 32 & 26 & 6 \\
\midrule
\textbf{Total} & & & \textbf{1,491} & \textbf{523} & \textbf{503} & \textbf{300} & \textbf{165} \\
\bottomrule
\end{longtable}
\endgroup
\twocolumn
\appendixspacing

\noindent\textit{Note.} The Total and D1/D2/D3/D4 columns report total claims
and the per-type breakdown.

The benchmark includes 1,491 SAU claims across 30 papers,
averaging 49.7 per paper (range 22--94).
D1 (numerical precision, 523 claims) and D2 (method/formula, 503 claims)
are the largest categories, followed by D3 (experimental protocol, 300
claims) and D4 (step ordering, 165 claims). The smallest paper
(\texttt{avg-reward-pg}, 22 claims) is an empirical RL study with a compact
method specification; the largest (\texttt{wdno}, 94 claims) describes a complex
neural operator architecture with extensive hyperparameter configurations.

%\vspace{-0.8\baselineskip}
\section{SAU Data Format and Examples}
\label{app:sau-examples}

\begin{table}[H]
\centering
\footnotesize
\setlength{\tabcolsep}{3pt}
\caption{SAU claims per dimension for the 30-paper benchmark.}
\label{tab:sau-counts}
\begin{tabular}{@{}lccccc@{}}
\toprule
\textbf{Statistic} & \textbf{D1} & \textbf{D2} & \textbf{D3} & \textbf{D4} & \textbf{Total} \\
\midrule
Total claims & 523 & 503 & 300 & 165 & 1,491 \\
\% of total & 35.1\% & 33.7\% & 20.1\% & 11.1\% & 100\% \\
Mean per paper & 17.4 & 16.8 & 10.0 & 5.5 & 49.7 \\
\bottomrule
\end{tabular}
\end{table}

Each SAU is stored in per-paper \texttt{sau.json} files with top-level arrays
keyed by drift type (\texttt{D1}--\texttt{D4}). Every claim has an \texttt{id}
(\texttt{\{paper\_id\}-D\{N\}-\{NNN\}}), a self-contained \texttt{claim} text,
and a \texttt{source} (paper evidence location). The format is minimal by design:
all information needed to judge correctness is in the claim text and source;
additional metadata is stored
separately to keep the benchmark annotation independent of the extraction
pipeline.

\subsection{Example: \texttt{nfig} SAU Annotation}
\label{app:sau-example}

We use \textit{NFIG: Multi-Scale Autoregressive Image Generation via Frequency
Ordering} (NeurIPS~2025) as a representative example.

\begingroup
\def\I#1{\hspace*{#1em}}
\begin{promptbox}{Listing 1: SAU annotation file structure --- paper \texttt{nfig}}{lst:nfig-sau}
\{\\
\I2"paper\_id": "nfig",\\
\I2"paper\_title": "NFIG: Multi-Scale Autoregressive Image Generation via Frequency Ordering",\\
\par\smallskip
\I2"D1": [\\
\I4\{...\}, \textit{\% 5 more D1 claims}\\
\I4\{\\
\I6"id": "nfig-D1-001",\\
\I6"claim": "ImageNet ILSVRC 2012: 1.2M train / 50k val /\\
\I8100k test images, 1000 categories, 256x256x3 resolution.",\\
\I6"source": "Section 4.1"\\
\I4\},\\
\I4\{\\
\I6"id": "nfig-D1-005",\\
\I6"claim": "Training: PyTorch, NVIDIA H100, Adam lr=8e-5,\\
\I8batch\_size=768, 350 epochs; inference: CFG=4.5, top\_k=990, 10 steps.",\\
\I6"source": "Section 4.1, Section 4.2"\\
\I4\}\\
\I2],\\
\par\smallskip
\I2"D2": [\\
\I4\{...\}, \textit{\% 8 more D2 claims}\\
\I4\{\\
\I6"id": "nfig-D2-003",\\
\I6"claim": "Residual Token Extraction: v\_0 = argmin\\
\I8||f\_hat\_0 - Z(v\_0)||\^{}2; R\_0 = f\_hat\_0 - Z(v\_0);\\
\I8for i>=1: v\_i = argmin ||(R\_\{i-1\}+f\_hat\_i) - Z(v\_i)||\^{}2;\\
\I8R\_i = R\_\{i-1\} + (f\_hat\_i - Z(v\_i)). Progressive\\
\I8frequency-band quantization with cumulative residual\\
\I8tracking unencoded signal through level i.",\\
\I6"source": "Section 3.1.2, Eq. 4"\\
\I4\},\\
\I2],\\
\par\smallskip
\I2"D3": [\\
\I4\{...\}, \textit{\% 6 more D3 claims}\\
\I4\{\\
\I6"id": "nfig-D3-006",\\
\I6"claim": "Frequency Keep Ability Analysis: compare NFIG vs\\
\I8VAR-16 on ImageNet using PSD (Power Spectral\\
\I8Density, lower is better) and FKS (Frequency Keep\\
\I8Score, weighted: Low 0.57, Mid 0.28, High 0.15;\\
\I8higher is better) across low/mid/high frequency bands.",\\
\I6"source": "Section 4.3"\\
\I4\}\\
\I2],\\
\par\smallskip
\I2"D4": [\\
\I4\{...\}, \textit{\% 6 more D4 claims}\\
\I4\{\\
\I6"id": "nfig-D4-001",\\
\I6"claim": "Pipeline: (1) Train FR-VAE tokenizer with\\
\I8frequency-guided residual quantization + VQGAN losses\\
\I8$\rightarrow$ (2) Train NFIG Transformer on FR-VAE tokens\\
\I8with cross-entropy loss $\rightarrow$ (3) Inference: generate\\
\I8tokens autoregressively from low to high frequency\\
\I8(10 steps), decode via FR-VAE decoder.",\\
\I6"source": "Section 4.1"\\
\I4\}\\
\I2]\\
\}
\end{promptbox}
\endgroup

% ===================================================================
%  APPENDIX C: RESULTS — Full evaluation tables
% ===================================================================

\FloatBarrier

\section{Pilot Study Detailed Results}
\label{app:pilot-study}

This appendix provides the full per-paper breakdown of the pilot study
summarized in Section~\ref{sec:design}. All runs used Claude Sonnet~4.6 with
BasicAgent (ReAct, max\_steps=80, time\_limit=900s) on five PaperBench-dev
papers, evaluated by the official PaperBench-dev pipeline with GPT-4o judge
(\texttt{code\_only=True}).

\subsection{Per-Paper Score Breakdown}

Table~\ref{tab:per-paper-detail} reports per-paper scores along with raw
resource usage harvested from each run's \texttt{meta.json} and
\texttt{grade.json} artifacts.

\begin{table*}[t]
\centering
\footnotesize
\setlength{\tabcolsep}{4pt}
\caption{Per-paper pilot results with resource usage.
Score is the gpt-4o judge's leaf-aggregated mean. Steps and time are read from
the agent's step timings; cost is the recorded API spend.}
\label{tab:per-paper-detail}
\begin{tabular}{@{}p{0.31\textwidth}rrrrrr@{}}
\toprule
\textbf{Paper} & \textbf{Score} & \textbf{Pass} & \textbf{Total} & \textbf{Steps} & \textbf{Time (s)} & \textbf{Cost} \\
\midrule
fre                       & 0.331 & 113 & 306 & 24 & 909.1  & \$2.11 \\
sample-specific-masks     & 0.603 & 66  & 87  & 81 & 652.4  & \$6.10 \\
mechanistic-understanding & 0.238 & 29  & 36  & 37 & 1252.5 & \$5.81 \\
pinn                      & 0.482 & 112 & 126 & 35 & 909.4  & \$4.10 \\
all-in-one                & 0.117 & 15  & 92  & 81 & 478.1  & \$3.41 \\
\midrule
\textbf{Total/Avg.}       & \textbf{0.581} & \textbf{335} & \textbf{647} & \textbf{52} & \textbf{840.3} & \textbf{\$21.53} \\
\bottomrule
\end{tabular}
\end{table*}

\subsection{Deriving the D1--D4 Taxonomy from Low-Score Nodes}
\label{app:pilot-low-score}

The four drift dimensions in Section~\ref{sec:design} were not pre-imposed:
they were summarized inductively from the failed rubric nodes in
Table~\ref{tab:per-paper-detail}. We took every leaf node receiving a score
of $0$ from the GPT-4o judge (312 nodes across the five papers) and read
the judge's natural-language verdict for each. We then grouped verdicts by
the kind of paper specification they reported as missing or wrong, without
fixing categories in advance.

Four recurring patterns emerged. Some verdicts identified
\textit{value-level} mismatches: a constant, threshold or layer width set
to the wrong number. Others identified \textit{algorithmic} substitutions
or omissions: the paper's loss, sampling step or attention mask was either
absent or replaced by an unrelated computation. A third group identified
\textit{protocol} gaps: the dataset, baseline or evaluation environment
named in the paper was simply not implemented. The remaining cluster
identified \textit{ordering} errors: pipeline phases run in the wrong order
or merged into a single pass. These four patterns became D1
(numerical precision), D2 (method/formula), D3 (experimental protocol) and
D4 (step ordering) in the main taxonomy. We applied no per-paper drift
labels: each paper contributes failures across multiple dimensions, and the
taxonomy is over individual claims, not whole reproductions.

\vspace{-0.8\baselineskip}
\subsection{Representative Judge Reasoning Excerpts}

Table~\ref{tab:judge-excerpts} shows one verbatim verdict per paper,
sampled from the 312 score-0 nodes analyzed in
\S\ref{app:pilot-low-score}.

\begin{table}[H]
\centering
\footnotesize
\caption{Verbatim excerpts from GPT-4o judge verdicts on score-0 leaf nodes
from the pilot runs. Quotations are abridged for length; ellipses mark
omitted text.}
\label{tab:judge-excerpts}
\begin{tabular}{@{}p{0.25\columnwidth}p{0.68\columnwidth}@{}}
\toprule
\textbf{Paper} & \textbf{Judge verdict (abridged, verbatim)} \\
\midrule
all-in-one & ``The submission lacks any explicit implementation of the
[Euler-Maruyama] discretization process, and no reference to the number
of steps (500) is observable in the provided files. The absence of such
code means the submission does not meet the resolution criteria.'' \\
\addlinespace
fre & ``The submission does not meet the resolution criteria. Although it
partially implements reward functions and describes the evaluation tasks
for the walker domain, there is no actual code to set up or interact with
the ExORL walker environment. The absence of integration with the
\texttt{custom\_dmc\_tasks} repository\ldots{} indicates a failure to
satisfy the criterion.'' \\
\addlinespace
sample-specific-masks & ``Instead of \texttt{32} output channels, the
third layer implements \texttt{64} output channels. The required
\texttt{BatchNorm} operation immediately following the third convolution
is missing.'' \\
\addlinespace
mechanistic-understanding & ``The code processes only a subset of 295
prompts instead of the required 1199 prompts from the
RealToxicityPrompts challenge set. There is no clear integration showing
measurement across all 20 token generations for every prompt.'' \\
\addlinespace
pinn & ``While it has implemented the ability to extract the \texttt{s},
\texttt{y}, and \texttt{rho} vectors with the function
\texttt{extract\_lbfgs\_state}\ldots{} it does not integrate this
functionality into the training pipeline to actually save these vectors at
the end of training. Without saving the L-BFGS state, the criterion is not
fully resolved.'' \\
\bottomrule
\end{tabular}
\end{table}

\section{SAU Extraction Architecture: Full Details}
\label{app:extraction-details}

This appendix provides the complete extraction architecture details summarized in
Section~\ref{sec:sa-bench}.

\subsection{Core Challenges}

Directly extracting code-implementable claims from ML papers with a single
LLM call faces two structural obstacles:

\begin{enumerate}[leftmargin=1.5em]
  \item \textbf{Attention degradation.} ML papers span 20--40 pages. A single
    LLM call processing the full text suffers severe attention decay,
    causing extraction quality to collapse in later sections, precisely
    where implementation-critical details (hyperparameter tables,
    pseudocode corrections) reside. The same effect produces a strong
    bias toward main-text scanning and skipping appendices, where the
    bulk of hyperparameter and protocol material lives.
  \item \textbf{Content-type entanglement.} Paper text mixes derivations,
    experimental results, baseline descriptions and notational bridge
    equations with the implementable claims. A general-purpose extraction
    prompt cannot reliably distinguish ``what the paper states'' from
    ``what the code needs.''
\end{enumerate}

\subsection{Agent-Per-Dimension Extraction}

We design three specialized extraction agents, each assigned to one
primary diagnostic dimension. Each agent is equipped with file-reading
and code-search tools. D4 is derived during the merge phase from ordering
annotations in D2/D3 outputs.

Each agent receives the full paper and internally spawns sub-agents per section
to parallelize deep reading. This avoids the attention degradation of
single-pass full-paper extraction: each sub-agent operates on 1--3 sections
where context quality is preserved and the parent agent synthesizes results.

\paragraph{Agent 1: Numerical Matcher (targeting D1).}
Scans the paper's section structure to identify numerically dense regions
(hyperparameter tables, configuration blocks, architecture descriptions), skips
primarily textual sections, spawns sub-agents in parallel each responsible for
1--3 sections. Each sub-agent performs regex pre-scanning for scientific
notation and unit-bearing numbers, classifies each hit using a decision tree
(filtering out result values, mathematical constants, proof/derivation values,
and numbers from cited prior work) and groups surviving values into
configuration groups. The parent agent deduplicates across sections.

\paragraph{Agent 2: Method Parser (targeting D2).}
Scans for method-related sections plus Appendix pseudocode and implementation
details. Spawns sub-agents per 1--2 sections for deep formula and algorithm
reading. Extracts \emph{all} code-implementable components (both novel
contributions and standard components) with complete variable definitions.
Records \texttt{ordering\_before} / \texttt{ordering\_after} annotations for
downstream D4 derivation. Limits each method component to 1--2 claims to avoid
formula fragmentation.

\paragraph{Agent 3: Protocol Enumerator (targeting D3).}
Scans for experiment-related sections plus Appendix experimental details.
Spawns sub-agents per 1--2 sections. Each extracted protocol binds four
elements: what is being compared, on what data, against what baselines and
with what metrics. Records \texttt{phase\_ordering} annotations. The parent
agent performs defensive merge validation: protocols must have full-sentence
descriptions, non-empty arrays and exceed 20 characters. If $\geq$30\% of
sub-agent outputs fail, the agent re-spawns with reduced section scope.

\subsection{D4 Derivation and Merge}

D4 claims are derived from ordering annotations in D2/D3 outputs during the
merge phase. Only ordering constraints \emph{explicitly stated} in the paper
qualify: numbered Step~1$\rightarrow$2$\rightarrow$3 in pseudocode, Phase~1
$\rightarrow$2$\rightarrow$3 training pipelines and explicit ``X before Y''
statements. Naturally implied orders (train then test) and code-engineering
call chains are excluded.

The merge coordinator performs: (1)~cross-agent deduplication; (2)~rule-based
classification (D1--D3 based on primary features); (3)~D4 derivation from
ordering annotations; (4)~LLM fallback for edge cases ($<$10\% of claims).

\subsection{Comparison with PaperBench Extraction}

\begin{table}[H]
\centering
\footnotesize
\caption{Comparison of SAU extraction methodology with PaperBench's
manually authored rubric trees.}
\label{tab:extraction-comparison-appendix}
\begin{tabular}{@{}p{0.24\columnwidth}p{0.32\columnwidth}p{0.32\columnwidth}@{}}
\toprule
\textbf{Dimension} & \textbf{PaperBench} & \textbf{Ours} \\
\midrule
Taxonomy            & Free-form rubric tree     & D1--D4 diagnostic \\
Extraction          & Author-paired manual      & Multi-agent section-level \\
Filtering           & Implicit (author judgment) & Code-first hard rules \\
Granularity control & Human rubric tree         & Rule-constrained $+$ human review \\
Effort per paper    & Author-time intensive     & Multi-agent parallel $+$ 1--2\,h human \\
\bottomrule
\end{tabular}
\end{table}

\subsection{Design Rationale}

This architecture addresses the two extraction challenges directly. The
sub-agent spawn pattern prevents attention degradation; specialized agents
with hard exclude rules outperform a single general-purpose prompt; and each
agent independently decides which sections to search, ensuring
implementation-critical details are captured regardless of paper location.
Despite code-first filtering, a mandatory human verification step (1--2
hours per paper) handles ambiguous cases around ablation hyperparameters and
derivation/implementation boundary formulas.

%\clearpage
\appendixspacing
\subsection{Sub-Agent Prompts}
\label{app:extraction-prompts}

The exact prompts dispatched to the D1/D2/D3 sub-agents are reproduced
below. Section assignment placeholders (\texttt{\{paper\_id\}},
\texttt{\{section\_list\}}, \texttt{\{start\}}, \texttt{\{end\}}) are
filled by the parent agent at dispatch time.

\vspace{0.2\baselineskip}
\noindent\textbf{D1 Numerical Matcher.}
\vspace{0.2\baselineskip}
\begin{promptbox}{Listing 2: D1 Numerical Matcher --- sub-agent prompt}{lst:prompt-d1}
You are a Numerical Matcher extracting code-relevant numerical values from
assigned sections of an ML paper.\\[2pt]
Paper: \{paper\_id\}; sections: \{section\_list\} (lines \{start\}-\{end\}).\\[4pt]
\#\# Filters\\
EXCLUDE: result tables, math constants ($\pi$, $e$, $\sqrt{2}$),
proof/derivation numbers, numbers from cited prior work, post-training
observed values.\\
INCLUDE: hyperparameters (lr, batch, $\tau$, $\lambda$), architecture
numbers (d\_model, n\_layers, dim), thresholds, scaling factors and data
numbers.\\[4pt]
\#\# Procedure\\
1.\ Grep-scan numeric patterns; 2.\ for each hit read $\pm 3$ lines and
classify; 3.\ group into config groups (e.g., "AdamW config").\\[4pt]
\#\# Output\\
JSON array of \{numerical\_value, parameter\_name, config\_group,
source\_section\}.
\end{promptbox}

\vspace{0.2\baselineskip}
\noindent\textbf{D2 Method Parser.}
\vspace{0.2\baselineskip}
\begin{promptbox}{Listing 3: D2 Method Parser --- sub-agent prompt}{lst:prompt-d2}
You are a Method Parser extracting code-implementable formulas and
algorithm components from assigned sections.\\[2pt]
Paper: \{paper\_id\}; sections: \{section\_list\}.\\[4pt]
\#\# What You Extract\\
Discrete computable formulas and algorithm steps that must appear in the
final codebase. Standard components (Transformer, AdamW, cross-entropy)
are valid claims if specified in the implementation.\\[4pt]
\#\# Hard Filters\\
EXCLUDE: theorems, lemmas, proofs, convergence bounds, continuous
ODE/SDE/PDE without a discrete solver, notation bridges, abstract claims,
result values.\\
INCLUDE: discrete iterative formulas, complete loss functions,
architecture components with structure, pseudocode steps from Algorithm
environments.\\[4pt]
\#\# Completeness\\
Each formula MUST be self-contained (define all symbols inline).\\[4pt]
\#\# Ordering\\
Record paper-specified ordering between components, not code-level call
chains or natural dependencies.\\[4pt]
\#\# Granularity\\
1--2 claims per named method unit (e.g., "Eigen Attention", "LERP Update").\\[4pt]
\#\# Output\\
JSON array of \{component, formula\_or\_algorithm, ordering\_before,
ordering\_after, source\_section\}.
\end{promptbox}

\vspace{0.2\baselineskip}
\noindent\textbf{D3 Protocol Enumerator.}
\vspace{0.2\baselineskip}
\begin{promptbox}{Listing 4: D3 Protocol Enumerator --- sub-agent prompt}{lst:prompt-d3}
You are a Protocol Enumerator extracting experiment protocols from
assigned sections.\\[2pt]
Paper: \{paper\_id\}; sections: \{section\_list\}.\\[4pt]
\#\# Definition\\
A valid protocol binds FOUR things: (1) Purpose, (2) Data, (3) Baselines,
(4) Metrics. Do NOT extract isolated dataset or metric names without
their protocol context.\\[4pt]
\#\# Hard Filters\\
EXCLUDE: isolated metric values, baseline descriptions of OTHER papers'
methods (unless reimplementation is required), ablation result numbers,
generic statements like "we evaluate on standard benchmarks."\\
INCLUDE: dataset names + train/val/test usage; baseline configurations;
per-protocol metrics; training setup (hardware, precision, framework);
data preprocessing pipelines.\\[4pt]
\#\# Phase Ordering\\
Record paper-specified multi-stage pipelines (pre-train $\to$ fine-tune
$\to$ evaluate); skip natural dependencies.\\[4pt]
\#\# Granularity\\
3--8 protocols per paper. Main experiments and ablations each get their
own entry.\\[4pt]
\#\# Output\\
JSON array of \{protocol\_description, datasets, baselines, metrics,
phase\_ordering, source\_section\}.
\end{promptbox}

\section{SAU Human Review Examples}
\label{app:review-examples}

Each paper's SAU extraction underwent mandatory author review. All review
actions are recorded in per-paper \texttt{review\_log.json} files with
action type, claim ID and reasoning. The multi-agent extraction pipeline
achieves high recall (it rarely misses a specification-relevant section)
but its precision is limited: agents cannot reliably distinguish paper
contributions from prior work or result values and misclassify D1--D4
types at a non-trivial rate. Consistent with the 17\% rejection rate
reported in Section~\ref{sec:sau-extraction}, the dominant review
actions are rejections and modifications, with smaller fractions of
merges, additions and type corrections. Below we provide representative
before/after examples organized by failure mode.

\subsection{Baseline Contamination}

The most consequential failure mode: the agent extracts methods from prior
work cited in the paper as if they were contributions. This is
fundamentally difficult because papers describe both prior methods and
their own methods using the same notation and terminology.

\vspace{0.2\baselineskip}
\begin{reviewbox}{Review Case 1: Baseline contamination --- paper \texttt{diffusion-convergence-rate}}{rev:baseline-contam}
\textbf{Paper:} diffusion-convergence-rate

\bigskip
\noindent\textbf{Agent extracted (DELETED):}
``The randomized midpoint sampler: for K rounds and N steps with $KN = 2T$,
sample from a randomized discretization scheme using learning rate
$\gamma_k = \ldots$ and noise injection $\sigma_k = \ldots$ (Eq.~8--11, Li
\& Jiao 2024).''

\noindent\textbf{Reviewer rationale:} ``The paper explicitly states `The
sampler used is the same as the one employed by Li and Jiao (2024)'
(Section~2.2). All four claims describing this sampler are not
contributions of the current paper (the paper contributes only the
analytical framework (theorem statements), not the algorithm.''

\noindent\textbf{Impact:} 6 of 12 agent-extracted claims deleted (50\%
rejection rate). The agent correctly identified all mathematical content
but could not discriminate paper contribution from cited prior work.
\end{reviewbox}

%\bigskip

\subsection{Result Value Extraction}

The agent extracts benchmark accuracy numbers as if they were
implementation specifications. This reflects a fundamental competency gap:
the agent does not understand that result values describe \emph{outcomes}
of the implemented system, not \emph{inputs} to it.

\vspace{0.2\baselineskip}
\begin{reviewbox}{Review Case 2: Result-value extraction --- paper \texttt{sam2}}{rev:result-value}
\textbf{Paper:} sam2

\bigskip
\noindent\textbf{Agent extracted (DELETED):}
``Semi-supervised VOS: 1-click achieves 64.7 J\&F, 3-click achieves 75.3
J\&F (Table~4). Image segmentation: 58.9 vs 58.1 mIoU, 6$\times$ faster,
61.9 mIoU on SA-23 (Table~5). VOS SOTA: 76.6 MOSE, 90.2 DAVIS (Table~6).
Prior methods peak at $\sim$60--62; SAM~2 achieves 76.8--78.4,
$+$15 point improvement.''

\noindent\textbf{Reviewer rationale:} ``Pure result values from Tables
4--6. Not actionable implementation claims. Evaluation protocol aspects
already captured in other D3 claims.''

\noindent\textbf{Impact:} A cluster of result-value claims rejected in
review. The agent systematically confuses descriptive result statements
with prescriptive implementation specifications.
\end{reviewbox}

\bigskip

\subsection{D1--D4 Type Misclassification}

The most frequent error: the agent misclassifies claims across the D1--D4
taxonomy. Architecture and workflow decisions (D4) are particularly
challenging: they are often mislabeled as D1 (numerical) or D2 (method).

\vspace{0.2\baselineskip}
\begin{reviewbox}{Review Case 3: D1$\leftrightarrow$D4 type misclassification --- paper \texttt{navil}}{rev:retype}
\textbf{Paper:} navil

\bigskip
\noindent\textbf{Agent extracted (RETYPED D1$\rightarrow$D4):}
``The visual encoder uses bidirectional attention with 2D-RoPE while the LLM
uses causal attention with 1D-RoPE (Section~3.1).''\\
\textbf{Reviewer:} ``Type changed D1$\rightarrow$D4. This is an
architecture design decision about attention mechanism and positional
encoding scheme, not a numerical hyperparameter value.''

\bigskip
\noindent\textbf{Agent extracted (RETYPED D1$\rightarrow$D4):}
``Observation~1: adding visual features before MoE injection is better than
adding after.''\\
\textbf{Reviewer:} ``Type changed D1$\rightarrow$D4. An empirical design
finding, not a formula or numerical specification.''

\bigskip
\noindent\textbf{Agent extracted (RETYPED D1$\rightarrow$D2):}
``Training stage: visual projector and LLM trained on image-text data with
language model loss, keeping vision encoder and connector frozen.''\\
\textbf{Reviewer:} ``Type changed D1$\rightarrow$D2. This is an
experimental protocol (training recipe) and the parameter freezing strategy
is a methodological choice, not a numerical configuration.''

\noindent\textbf{Impact:} 10 of 35 agent-extracted claims in navil were
modified, predominantly to correct D1--D4 type assignments (29\% revision
rate; see Table~\ref{tab:review-stats}). The pattern is systematic across
all papers: agents default to D1 for any claim containing a number or
named component and struggle to identify D4 ordering constraints.
\end{reviewbox}

\bigskip

\subsection{Over-Splitting and Granularity Errors}

Agents frequently fragment a single implementation component into multiple
claims, each describing a sub-component that has no independent code
existence.

\vspace{0.2\baselineskip}
\begin{reviewbox}{Review Case 4: Over-splitting and granularity errors --- papers \texttt{navil}, \texttt{conformal-bayesian-quadrature}}{rev:over-split}
\textbf{Paper:} navil

\bigskip
\noindent\textbf{Agent extracted (MERGED, 3$\rightarrow$1):}
\begin{itemize}[nosep]
  \item req-005: ``MHA-MMoE: visual token Attention with modality-specific
    QKV projections and modality indicator embeddings.''
  \item req-006: ``FFN-MMoE: visual token FFN with modality-specific MLP
    experts and indicator embeddings.''
  \item req-007: ``Single-expert activation: only the top-1 expert is
    activated per token in both attention and FFN MoE layers.''
\end{itemize}

\noindent\textbf{Reviewer:} ``All three claims describe the same
modality-specific MoE architecture in Section~3.2.2. MHA-MMoE, FFN-MMoE,
and single-expert activation are inseparable sub-components of one MoE
design (they would be implemented together in the same code module). Merged
into a single D1 claim.''

\bigskip
\noindent\textbf{Paper:} conformal-bayesian-quadrature

\bigskip
\noindent\textbf{Agent extracted (MERGED, 3$\rightarrow$1):}
\begin{itemize}[nosep]
  \item req-006: ``HPD decision rule formula (Section~4.1).''
  \item req-017: ``The 7-step algorithm pipeline for computing the HPD set.''
  \item req-018: ``$\lambda$ search procedure (grid/line search).''
\end{itemize}

\noindent\textbf{Reviewer:} ``All three describe the same core method,
the HPD decision rule and its implementation, at different granularities. The
pipeline is the direct implementation of the formula; the $\lambda$ search
is step~(7) of the same pipeline. Merged into a single D1 claim.''
\end{reviewbox}

\bigskip

\subsection{Omissions: What the Agent Misses}

Agents also exhibit false negatives: specifications that appear deep in
appendices, are stated purely in prose, or rely on notational conventions
are systematically missed. A second source of false negatives is structural:
the auto-extraction prompt for D2 conservatively excludes pure theorems,
lemmas and convergence bounds (Listing~\ref{lst:prompt-d2}) to keep
SAUs implementation-grounded. When such a result is the paper's primary
contribution and is operationally re-used elsewhere in the pipeline, the
human reviewer overrides this filter and adds the SAU back, attaching the
relevant downstream code-implementable surface so it remains scorable. Case
5 illustrates both patterns.

\vspace{0.2\baselineskip}
\begin{reviewbox}{Review Case 5: Omissions --- papers \texttt{diffusion-convergence-rate}, \texttt{conformal-bayesian-quadrature}}{rev:omissions}
\textbf{Paper:} diffusion-convergence-rate (appendix-buried protocol)

\smallskip
\noindent\textbf{Missing (ADDED by reviewer):}
``Iteration budget $K = c_2 \min\{d \log^2 T,\; L \log T\}$ for the
discrete sampler (Appendix~B). This schedule is the operational form of
the headline complexity bound and must be set explicitly in the training
loop; the agent extracted the sampler step but omitted the rule for
choosing $K$, leaving it as a free hyperparameter.''

\smallskip
\noindent\textbf{Reviewer:} ``The headline theorem itself is excluded by
the D2 prompt, but its operational consequence---the iteration schedule
the implementation must follow---is a concrete D2 claim that was
missed.''

\smallskip
\noindent\textbf{Paper:} conformal-bayesian-quadrature (override of D2 filter)

\smallskip
\noindent\textbf{Missing (ADDED by reviewer):}
``Construction of $L^+$ via the worst-case quantile bound
$\sup_\pi \mathbb{E}[L \mid t, \ell] \le \sum_i u_i \ell_{(i)}$
(Section~4.3). The bound itself is theorem-shaped, but the algorithm
returns $L^+$ directly built from it, so the human reviewer attaches
the formula to the constructive code path and admits it as D2.''
\end{reviewbox}

\noindent Across all 30 papers, the most frequently missed specification
type is appendix-only protocol details (agents underweight appendix content
despite explicit instructions) and operational consequences of theoretical
results (agents extract the surrounding implementation but skip the
schedule, threshold or constant the theorem prescribes).

\appendixspacing

\subsection{Per-Paper Review Summary}

The full review log for each paper (with per-claim before/after text and
reviewer reasoning) is included in the benchmark release. The extraction
pipeline and judge prompts are model-agnostic; we encourage future work to
test alternative extraction backends and cross-institutional validation
of the scoring protocol.

\FloatBarrier

\section{Judge Prompt Templates}
\label{app:judge-prompts}

The following prompts are used by the SAU scoring judge. Each claim is evaluated
with dimension-aware prompts calibrated by the per-type scoring rubric
(Appendix~\ref{app:scoring-rubric}).

\subsection{System Prompt}

\begin{promptbox}{Listing 5: Judge System Prompt --- five-level scoring rubric}{lst:judge-system}
You are a strict SAU code-reproduction judge.\\
Return JSON only. Score must be exactly one of: 0, 0.25, 0.5, 0.75, 1.\\
\par\smallskip
Judge using these rules:\\
- Evidence from code/config matters most; docs/tests may only corroborate.\\
- Score the implementation, not the paper text or the repo README.\\
- Prefer lower scores unless the claim's concrete mechanism is visible in code/config.\\
- If the evidence mostly shows placeholders, stubs, comments, or toy code, keep the score low.\\
- Return concise reasoning with file:line references only for the strongest matches.\\
- Every result object must include claim\_id, score, reasoning and evidence\_refs.\\
- Do not omit fields and do not return null for any required field.\\
- Write reasoning as plain text only. Do not use LaTeX commands, backslashes, or unescaped math markup.
\end{promptbox}

\subsection{Single-Claim User Prompt}

The judge receives a JSON payload with the SAU claim, its dimension type
(D1--D4), paper source spans and the top-2 direct evidence items retrieved
from the agent's repository. The expected output schema enforces structured
scoring with cited evidence.

\begin{promptbox}{Listing 6: Single-Claim User Prompt --- example payload (\texttt{emergent-planning-rl} D1-004)}{lst:judge-single}
\{\\
~~"claim\_id": "emergent-planning-rl-D1-004",\\
~~"dimension": "D1",\\
~~"claim": "DRC(3,3) agent trained via IMPALA on 900,000 Boxoban\\
~~~~~~~levels for 250M transitions, with discount factor $\gamma{=}0.97$.",\\
~~"source": "Appendix E.4",\\
~~"direct\_evidence": [\\
~~~~~\{ "file": "configs/drc33.yaml",\\
~~~~~~~"lines": "109-116",\\
~~~~~~~"snippet": "n\_train\_episodes: 3000 ... IMPALA ... boxoban" \},\\
~~~~~\{ "file": "agents/drc\_agent.py",\\
~~~~~~~"lines": "77-80",\\
~~~~~~~"snippet": "discount = 0.99 ... transitions = 250\_000\_000" \}\\
~~],\\
~~"search\_trace": [\\
~~~~~\{ "query": "IMPALA 250M transitions discount",\\
~~~~~~~"hits": 3, "files\_searched": ["configs/", "agents/"] \}\\
~~],\\
~~"expected\_output": \{\\
~~~~~"claim\_id": "emergent-planning-rl-D1-004",\\
~~~~~"score": "0|0.25|0.5|0.75|1",\\
~~~~~"reasoning": "concise explanation with file:line refs",\\
~~~~~"evidence\_refs": ["path:line-or-range"]\\
~~\},\\
~~"required\_fields": ["claim\_id", "score", "reasoning", "evidence\_refs"]\\
\}
\end{promptbox}

\subsection{Batch Evaluation Format}

For efficiency, up to 4 claims are evaluated in a single LLM call. The batch
prompt wraps multiple claims in a \texttt{\{"claims": [\ldots]\}} envelope and
expects a \texttt{\{"results": [\ldots]\}} response with one score per claim.

\vspace{0.2\baselineskip}
\begin{promptbox}{Listing 7: Batch Evaluation Format --- up to 4 claims per LLM call}{lst:judge-batch}
\{\\
~~"claims": [\\
~~~~~\{ "claim\_id": "...", "dimension": "D1", "claim": "...",\\
~~~~~~~"source": "...", "direct\_evidence": [...], "search\_trace": [...] \},\\
~~~~~\{ "claim\_id": "...", "dimension": "D2", "claim": "...",\\
~~~~~~~"source": "...", "direct\_evidence": [...], "search\_trace": [...] \}\\
~~],\\
~~"expected\_output": \{\\
~~~~~"results": [\\
~~~~~~~\{ "claim\_id": "...", "score": "0|0.25|0.5|0.75|1",\\
~~~~~~~~~"reasoning": "concise explanation with file:line refs",\\
~~~~~~~~~"evidence\_refs": ["path:line-or-range"] \}\\
~~~~~],\\
~~~~~"required\_fields": ["claim\_id", "score", "reasoning", "evidence\_refs"]\\
~~\}\\
\}
\end{promptbox}

\noindent Dimension-specific scoring context is provided per claim via the
\texttt{dimension} field; the five-level rubric
(Appendix~\ref{app:scoring-rubric}) is loaded into the system prompt. The
zero-score pre-filter (Section~\ref{sec:judge-scoring}) eliminates
$\sim$30--40\% of claims without an LLM call via five heuristic rules: claim
keyword absent, search returns empty, repo size below threshold, only
docs/tests evidence and experiment-parameter claims lacking numeric support.

\section{Full Scoring Rubric Table}
\label{app:scoring-rubric}

Table~\ref{tab:scoring-rubric-full} provides the complete drift-type-specific
scoring rubric loaded into the judge prompt. The five-level semantic scale
(Section~\ref{sec:task-formulation}) is calibrated per drift type because
deviation severity has different practical consequences across dimensions.

\paragraph{Annotated scoring examples.}
The following examples (Table~\ref{tab:scoring-examples}) are loaded into the
judge prompt as calibration references. They are drawn from real pilot study
cases and cover all four drift types.

\begin{table*}[h!]
\centering
\footnotesize
\caption{Representative per-paper review statistics, drawn from
\texttt{review\_log.json}. \texttt{Before} = agent-extracted candidates
submitted for review (modifiable subset); \texttt{After} = candidates
after review. \texttt{Del/Mod/Add/Merge} are explicit reviewer actions on
the modifiable subset. The five papers illustrate different dominant
failure modes.}
\label{tab:review-stats}
\setlength{\tabcolsep}{3pt}
\begin{tabular}{@{}p{0.28\textwidth}rrrrrrp{0.24\textwidth}@{}}
\toprule
\textbf{Paper} & \textbf{Before} & \textbf{After} & \textbf{Del} &
\textbf{Mod} & \textbf{Add} & \textbf{Merge} & \textbf{Dominant Issue} \\
\midrule
diffusion-convergence-rate     & 12 & 6  & 6 & 1  & 2 & 2 & Baseline contamination \\
conformal-bayesian-quadrature  & 18 & 15 & 0 & 0  & 1 & 5 & Over-splitting \\
navil                          & 35 & 31 & 2 & 10 & 1 & 1 & Type misclassification \\
luno                           & 19 & 25 & 0 & 9  & 5 & 0 & Omissions, source fixes \\
moe-pot                        & 32 & 35 & 7 & 13 & 10 & 0 & Mixed: result values $+$ retypes \\
\bottomrule
\end{tabular}

\vspace{0.5\baselineskip}

\caption{Full five-level scoring rubric by drift type.}
\label{tab:scoring-rubric-full}
\begingroup
\footnotesize
\setlength{\tabcolsep}{2pt}
\renewcommand{\arraystretch}{0.90}
\begin{tabular}{@{}c|p{0.22\textwidth}p{0.22\textwidth}p{0.25\textwidth}p{0.22\textwidth}@{}}
\toprule
\textbf{Score} & \textbf{D1: Numerical Precision} & \textbf{D2: Method / Formula} & \textbf{D3: Experimental Protocol} & \textbf{D4: Step Ordering} \\
\midrule
1.0 & All parameter values match the paper exactly & All algorithm steps present and correctly implemented; formula transcription accurate & All baselines, datasets, metrics and ablations present and correctly configured & All paper-specified execution order constraints correctly implemented \\
0.75 & Order-of-magnitude consistent (e.g., lr $0.001\!\rightarrow\!0.003$), or optimizer-equivalent substitution & Core algorithm correct; minor organizational differences that do not alter semantics & Primary elements present; $\le 1$ non-core baseline or metric missing & Minor ordering variation that preserves dependencies (e.g., parallelizable steps reordered) \\
0.5 & Specific value deviates $>\!3\times$ but order-of-magnitude correct; or multiple minor parameters missing & Core algorithm present but missing $\ge 1$ key step (e.g., 4/5 steps), or conditional logic omitted & $\ge 2$ baselines missing, or a key dataset/metric absent & Key phase sequence constraint violated (e.g., 2 of 3 phases correctly ordered) \\
0.25 & Order-of-magnitude error (e.g., lr $0.001\!\rightarrow\!0.1$), or all hyperparameters at defaults & Core mechanism replaced by a fundamentally different method family (e.g., score matching $\rightarrow$ KL divergence) & Only $\ge 1$ related experimental elements present; majority absent & Core ordering reversed (e.g., fine-tuning before pre-training) \\
0 & All numerical values missing or incorrect & Implementation entirely missing or logic contradicts the paper & No experimental infrastructure present & No ordering structure; all phases merged into one \\
\bottomrule
\end{tabular}
\endgroup
\end{table*}

\begin{table*}[h!]
\centering

\caption{Annotated scoring examples for judge calibration.}
\label{tab:scoring-examples}
\begingroup
\footnotesize
\setlength{\tabcolsep}{2.5pt}
\renewcommand{\arraystretch}{0.92}
\begin{tabular}{@{}p{2.9cm}p{3.05cm}cp{0.75cm}p{5.75cm}@{}}
\toprule
\textbf{SAU Claim} & \textbf{Repo Implementation} & \textbf{Type} & \textbf{Score} & \textbf{Rationale} \\
\midrule
``Adam optimizer, lr=0.001'' & \texttt{AdamW(lr=0.001)} & D1 & 0.75 & Optimization variant: Adam$\rightarrow$AdamW is a functional improvement; does not alter experimental conclusions. \\
``Adam optimizer, lr=0.001'' & \texttt{SGD(lr=0.1)} & D1 & 0.25 & Core violation: different optimizer family + order-of-magnitude lr error; training dynamics fundamentally altered. \\
``Adam optimizer, lr=0.001'' & \texttt{Adam(lr=0.003)} & D1 & 0.75 & Minor deviation: order-of-magnitude consistent; specific value within $3\times$. \\
``Score matching loss (Eq.~5)'' & Code implements KL divergence & D2 & 0.25 & Core violation: different algorithm family; paper's core claim no longer holds. \\
``Score matching loss (Eq.~5)'' & Correct score matching but missing $\varepsilon$ regularization term & D2 & 0.5 & Key omission: core algorithm present but incomplete. \\
``Compare against PPO, SAC, TD3'' & Only PPO and SAC implemented & D3 & 0.5 & Key omission: 1 of 3 baselines missing. \\
``Compare against PPO, SAC, TD3'' & No baselines implemented & D3 & 0 & Complete absence. \\
``Phase 1 pre-training $\rightarrow$ Phase 2 fine-tuning'' & Both phases merged into single training stage & D4 & 0.25 & Core violation: paper-specified phase ordering collapsed; independent phase verification impossible. \\
\bottomrule
\end{tabular}
\endgroup

\vspace{0.25\baselineskip}

\caption{Scaffold hyperparameters and tool inventories for the three
generator scaffolds. Per-paper wall-clock cap is 2700\,s for all runs.}
\label{tab:scaffold-config}
\begingroup
\footnotesize
\setlength{\tabcolsep}{2pt}
\renewcommand{\arraystretch}{0.98}
\begin{tabular}{@{}p{0.16\textwidth}p{0.78\textwidth}@{}}
\toprule
\textbf{Scaffold} & \textbf{Configuration} \\
\midrule
BasicAgent  &
  ReAct loop; \texttt{max\_steps}=150;
  per-call LLM \texttt{timeout}=300\,s;
  tools: \texttt{bash}, \texttt{read\_file\_chunk},
  \texttt{search\_file} (local ripgrep), \texttt{submit};
  no web/network access. \\
\midrule
OpenHands   &
  CodeActAgent in local Docker runtime;
  \texttt{max\_iterations}=100;
  per-call LLM \texttt{timeout}=900\,s;
  enabled tools: shell, file editor, think, finish;
  jupyter, browser and MCP disabled. \\
\midrule
PaperCoder  &
  3-stage pipeline (planning $\rightarrow$ analyzing $\rightarrow$
  coding); per-stage \texttt{timeout}=3600\,s; templates inject explicit
  paper-parsing inductive bias. \\
\bottomrule
\end{tabular}
\endgroup

\vspace{0.25\baselineskip}

\caption{Per-paradigm mean SAU scores across all 12 generator configurations.}
\label{app:paradigm-scores}
\begingroup
\footnotesize
\setlength{\tabcolsep}{4pt}
\begin{tabular}{@{}lcccccc@{}}
\toprule
\textbf{Paradigm} & \textbf{Papers} & \textbf{Overall} & \textbf{D1} & \textbf{D2} & \textbf{D3} & \textbf{D4} \\
\midrule
New Algorithm / Architecture & 13 & 0.232 & 0.307 & 0.269 & 0.164 & 0.188 \\
Theoretical Analysis         & 5  & 0.225 & 0.276 & 0.222 & 0.192 & 0.211 \\
System / Pipeline            & 3  & 0.217 & 0.269 & 0.231 & 0.163 & 0.203 \\
Empirical Comparison         & 2  & 0.212 & 0.239 & 0.290 & 0.145 & 0.173 \\
Generative Models            & 2  & 0.210 & 0.351 & 0.190 & 0.125 & 0.172 \\
Incremental Improvement      & 5  & 0.198 & 0.268 & 0.210 & 0.135 & 0.179 \\
\bottomrule
\end{tabular}
\endgroup
\end{table*}

\subsection{Complete Judge Output Format}
\label{app:judge-output-schema}

Figure~\ref{fig:sau-score} in the main text shows a representative scoring output.
Below is the complete JSON schema for a single scored claim, using a real D2 claim
from the \texttt{adjoint-matching} paper (Claude~+~BasicAgent). The output includes
the claim identifier, the score, up to 8 evidence items with file:line references
and relevance tags, the search trace and \texttt{judge\_reasoning} (the
natural-language explanation that substantiates the score and enables downstream
error taxonomy, Section~\ref{sec:results-error}).

\def\I#1{\hspace*{#1em}}
\begin{singlejsonbox}{Listing 8: Complete Judge Output Schema --- paper \texttt{adjoint-matching}, claim D2-001 (Claude~+~BasicAgent)}{lst:judge-output}
\{\\
\I2"id": "adjoint-matching-D2-001",\\
\I2"claim": "Unified SDE: dX\_t = b(X\_t,t)dt + \ensuremath{\sigma}(t)dB\_t,\\
\I4b(x,t)=\ensuremath{\kappa}\_t\textperiodcentered x + (\ensuremath{\sigma}(t)\^{}2/2 + \ensuremath{\eta}\_t)\textperiodcentered\ensuremath{\mathfrak{s}}(x,t),\\
\I4\ensuremath{\kappa}\_t=\ensuremath{\dot{\alpha}}\_t/\ensuremath{\alpha}\_t,\\
\I4\ensuremath{\eta}\_t=\ensuremath{\beta}\_t(\ensuremath{\dot{\alpha}}\_t/\ensuremath{\alpha}\_t\textperiodcentered\ensuremath{\beta}\_t - \ensuremath{\dot{\beta}}\_t)",\\
\I2"source": "Section 3, Eq. 10-11",\\
\I2"score": 0.25,\\
\I2"search\_rounds": 3,\\
\I2"evidence": [\\
\I4\{\\
\I6"file": "src/sde\_simulation.py",\\
\I6"lines": "7-14",\\
\I6"snippet": "The unified SDE form (Eq. 10-11): dX\_t = b(X\_t, t)dt...",\\
\I6"relevance": "direct",\\
\I6"kind": "code"\\
\I4\},\\
\I4\{\\
\I6"file": "src/noise\_schedules.py",\\
\I6"lines": "123-130",\\
\I6"snippet": "sigma\_t = sqrt(2 * eta\_t) where eta\_t = beta\_t * (alpha\_dot\_t / alpha\_t)...",\\
\I6"relevance": "direct",\\
\I6"kind": "code"\\
\I4\},\\
\I4\textit{\% 6 more evidence items}\\
\I2],\\
\I2"judge\_reasoning": "The code computes \ensuremath{\eta}\_t and \ensuremath{\sigma}(t)\\
\I4in noise\_schedule.py, but there is no evidence that these\\
\I4are assembled into the full SDE drift b(x,t). The evaluator\\
\I4uses a different parameterization (velocity for the deterministic\\
\I4ODE), which does not implement the claimed stochastic drift.\\
\I4Only a partial match: formulas for components exist but the\\
\I4core SDE mechanism is missing.",\\
\I2"search\_trace": [\textit{\% 29 search steps omitted for brevity}]\\
\}
\end{singlejsonbox}

\noindent\textbf{Key design points.} The \texttt{evidence} array gives the judge (and
readers) file:line references for every alignment judgment, making the scoring
auditable. The \texttt{judge\_reasoning} field separately documents \emph{why} the
score was assigned, identifying what is present, what is missing and what is
incorrect. This separation of evidence and reasoning is what enables systematic
error taxonomy without additional human annotation: the reasoning text itself is
the annotation.

\appendixspacing

% ===================================================================
%  APPENDIX A.5: SCAFFOLD CONFIGURATIONS
% ===================================================================

\FloatBarrier
\section{Scaffold Configurations}
\label{app:scaffold-config}

Table~\ref{tab:scaffold-config} lists the per-scaffold hyperparameters used for
all 360 evaluation runs in Section~\ref{sec:evaluation-setup}. All four models
share identical scaffold settings; only the underlying LLM endpoint differs.

All scaffolds operate on the same source paper (\texttt{paper.md} produced
by MinerU v2.7.6~\cite{mineru}) and write to per-run workspaces under
\texttt{experiments/runs/\{model\}\_\{scaffold\}/}. BasicAgent and OpenHands
consume the markdown directly; PaperCoder requires S2ORC-format JSON~\cite{s2orc}, so
each \texttt{paper.md} is converted to \texttt{\{paper\_id\}.json} via
\texttt{convert\_md\_to\_s2orc.py} before the pipeline is invoked. Network
access during code generation is restricted to model-API calls; no
generator may fetch external repositories. The 2700\,s per-paper cap is
enforced by the batch runner regardless of scaffold-internal step or
iteration limits.

% ===================================================================
%  APPENDIX B: DATA — Benchmark composition and SAU examples
% ===================================================================

%\clearpage

\section{Full Results Tables}
\label{app:full-results}

Table~\ref{tab:dimension-stats-full} reports detailed per-dimension
statistics aggregated across all 12 generators and 30 papers. The
per-configuration scores referenced in Section~\ref{sec:results} are
provided in the main text (Table~\ref{tab:generator-scores-full}).

\begin{table*}[t!]
\centering
\begin{minipage}[t]{0.44\textwidth}
\centering
\footnotesize
\caption{Per-dimension statistics across all 12 generators and 30 papers.}
\label{tab:dimension-stats-full}
\setlength{\tabcolsep}{2.5pt}
\begin{tabular}{@{}lrrrr@{}}
\toprule
\textbf{Statistic} & \textbf{D1} & \textbf{D2} & \textbf{D3} & \textbf{D4} \\
\midrule
Mean                  & 0.290 & 0.244 & 0.160 & 0.190 \\
Median                & 0.304 & 0.235 & 0.167 & 0.188 \\
\% $\geq$0.5          & 28.7\% & 18.2\% & 3.7\% & 5.9\% \\
Full-mark rate (1.0)  & 9.7\% & 5.4\% & 0.7\% & 0.9\% \\
Zero rate             & 39.6\% & 35.8\% & 44.8\% & 33.7\% \\
\bottomrule
\end{tabular}
\end{minipage}
\hfill
\begin{minipage}[t]{0.50\textwidth}
\centering
\footnotesize

\caption{Top-5 easiest and bottom-5 hardest papers by mean SAS.}
\label{tab:paper-difficulty-app}
\begin{tabular}{@{}llc@{}}
\toprule
\textbf{Rank} & \textbf{Paper} & \textbf{Mean SAS} \\
\midrule
\multicolumn{3}{c}{\textit{Easiest}} \\
1 & conformal-bayesian-quadrature   & 0.304 \\
2 & mrq                             & 0.271 \\
3 & olmoe                           & 0.260 \\
4 & sc-fno                          & 0.257 \\
5 & neural-operator-flow-matching-pde & 0.255 \\
\midrule
\multicolumn{3}{c}{\textit{Hardest}} \\
26 & navil                         & 0.179 \\
27 & cara                          & 0.176 \\
28 & pyramidal-flow-matching       & 0.171 \\
29 & prioritized-generative-replay & 0.158 \\
30 & sam2                          & 0.158 \\
\bottomrule
\end{tabular}
\end{minipage}

\end{table*}

\begin{figure}[H]
\centering
\includegraphics[width=\columnwidth]{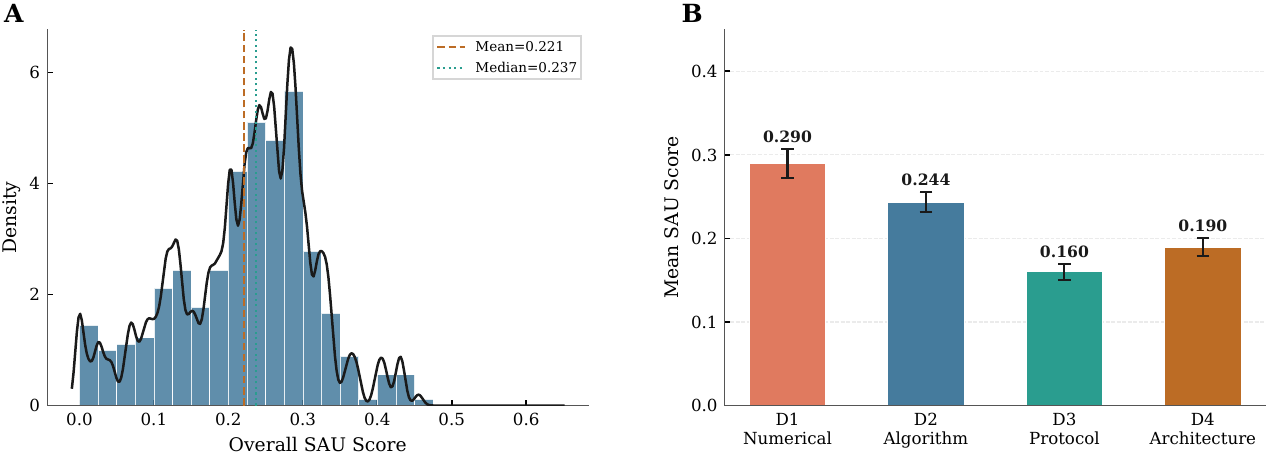}
\caption{Score distribution across all 360 evaluations. The heavy right skew
confirms semantic drift is pervasive.}
\label{fig:score-dist-app}
\end{figure}

\section{Case Analysis: Representative Failures}
\label{app:case-analysis}

The complete six-type failure taxonomy for zero-scored claims:
\begin{enumerate}[leftmargin=1.5em, nosep]
  \item \textbf{Implementation mismatch (40.8\%).} The judge located code referencing the claim's keywords, but determined the implementation does not match the specified content: the agent wrote something unrelated under recognizable names.
  \item \textbf{Stub/placeholder (16.2\%).} Evidence exists but is a stub, TODO, or \texttt{pass} statement.
  \item \textbf{External knowledge gap (8.0\%).} The claim requires implementing a standard baseline like PPO or loading a benchmark dataset like MOSE, which the paper names but does not define.
  \item \textbf{Token mention only (3.7\%).} Claim keyword appears in comments, docstrings, or imports, but no functional implementation exists.
  \item \textbf{Wrong implementation ($<$1\%).} Code references the specification but implements it incorrectly (e.g., optimizer substitution, formula error).
  \item \textbf{Missing component ($<$1\%).} Partial implementation present but a key sub-component is absent.
\end{enumerate}
The remaining $\sim$30\% of zero-scored claims resist single-category classification (multiple co-occurring failure modes or ambiguous reasoning).

\begin{table*}[t!]
\centering
\footnotesize
\caption{Representative failure cases from Claude + PaperCoder (best
configuration), annotated with failure taxonomy types.}
\label{tab:bad-cases}
\setlength{\tabcolsep}{2pt}
\renewcommand{\arraystretch}{0.92}
\begin{tabular}{@{}p{0.72cm}p{5.85cm}p{5.85cm}p{2.43cm}@{}}
\toprule
\textbf{Dim.} & \textbf{Paper Specification} & \textbf{Agent Output / Root Cause} & \textbf{Failure Type} \\
\midrule
D1 & Training: 900K levels, 250M transitions;
UNet: ch\_mult=[1,2,4,8] &
Uses 10K levels, ch\_mult=[1,1,2,4].
Specs in appendix, defaults not overridden & Impl. mismatch \\
\midrule
D2 & Adjoint matching loss with correction term;
SCoRe: two-turn RL with asymmetric reward &
Standard flow matching without correction; single-turn RL.
Mathematical distinctions missed & Wrong impl. \\
\midrule
D3 & 17 datasets (DAVIS, YT-VOS, MOSE, etc.);
5 baselines (PPO, SAC, TD3, Dreamer-v3, MPC) &
Data loading for 3 datasets; only PPO implemented & External knowledge \\
\midrule
D4 & 4-phase pipeline;
warmup$\rightarrow$cosine decay$\rightarrow$eval &
All phases merged; flat LR schedule.
Training recipe as prose & Ordering collapse \\
\bottomrule
\end{tabular}
\end{table*}

We present four representative failure cases from Claude-Sonnet-4.6 with
OpenHands on the \texttt{sam2} paper (77 SAU claims), one per drift
dimension, illustrating how the D1--D4 taxonomy enables fine-grained
diagnosis beyond aggregate scores.

\subsection{Case 1: Numerical Precision (D1) --- Image Encoder Architecture}

\textbf{Paper specification (D1-003).} SAM~2 uses an MAE-pre-trained
Hiera-B+ image encoder with embedding dimensions {[}96, 192, 384, 768{]}
across four stages, 12 attention heads, window size 14--28 per stage,
and global attention at the last stage. The encoder processes
$1024\times 1024$ input images through overlapping patch embedding (stride 4).

\textbf{Agent implementation.} The agent substituted a standard ViT-B/16
backbone (embed\_dim=768, depth=12, num\_heads=12, patch\_size=16). The
Hiera architecture (hierarchical stages with varying spatial resolution
and window-based attention) was replaced by uniform ViT blocks.
MAE pre-training initialization was omitted.

\textbf{Root cause.} \textit{Specification misreading.} The agent defaulted
to a well-known ViT implementation rather than implementing the paper's
Hiera architecture.

\subsection{Case 2: Method / Formula (D2) --- Memory Attention Mechanism}

\textbf{Paper specification (D2-007).} The memory attention module
stores spatial features $\mathbf{F} \in \mathbb{R}^{B\times N\times C}$
for $N=6$ past frames with $C=256$ channels and computes pointwise
cross-attention at every spatial location:
$\mathbf{A} = \text{softmax}(\mathbf{Q}\mathbf{K}^T / \sqrt{d_k})$,
$\mathbf{O} = \mathbf{A}\mathbf{V}$, followed by a residual
connection and an output linear projection.

\textbf{Agent implementation.} The agent implemented temporal
concatenation without cross-attention: past-frame features are flattened
and concatenated to the current frame, then passed through a standard
self-attention block. The pointwise spatial memory attention is absent;
the agent's code performs global attention instead.

\textbf{Root cause.} \textit{Core mechanism substitution.} The agent
substituted the paper's specialized memory cross-attention with a generic
temporal concatenation + self-attention pattern.

\subsection{Case 3: Experimental Protocol (D3) --- Evaluation Setup}

\textbf{Paper specification (D3-004--D3-009).} SAM~2 is evaluated on
video object segmentation across 9 datasets (DAVIS 2017, YouTube-VOS
2018/2019, MOSE, BURST, LVOS, SA-V, LV-VIS, UVO), each with
dataset-specific protocols (J \& F for DAVIS,
J\_seen, F\_seen, J\_unseen, F\_unseen for YouTube-VOS), plus 15
zero-shot video instance segmentation datasets and interactive
segmentation benchmarks.

\textbf{Agent implementation.} The agent produced evaluation scripts for
only 2 of 9 VOS datasets (DAVIS 2017 and a skeleton for YouTube-VOS),
with placeholder TODOs for the remaining 7. Zero-shot VIS and interactive
segmentation benchmarks are entirely absent.

\textbf{Root cause.} \textit{External knowledge gap.} Dataset integration
requires per-dataset downloading, preprocessing and protocol implementation.
The agent lacks external knowledge of dataset formats and treats evaluation
as a post-processing step rather than a primary deliverable.

\subsection{Case 4: Step Ordering (D4) --- Three-Phase Data Engine Collapse}

\textbf{Paper specification (\texttt{sam2-D4-002}).} The SA-V data engine
is a three-phase, model-in-the-loop annotation pipeline: \emph{Phase~1}
human annotators label every frame with pixel-perfect masks using SAM~2
interactive mode (high quality, slow); \emph{Phase~2} SAM~2 propagates
masks across frames and annotators only correct errors (medium quality,
6$\times$ faster); \emph{Phase~3} fully automatic masklet generation with
SAM~2 followed by quality filtering (low cost, scaled). Phases must be
executed in order: each later phase consumes data and a model trained on
the previous phase's output.

\textbf{Agent implementation.} Across all three Claude scaffolds the
generated repository contains only a flat \texttt{video\_dataset.py}
loader that reads pre-existing SA-V masklets. None of the three phases
appears as a callable stage; there is no human-annotation entry point,
no propagate-and-correct loop wrapping the model, and no automatic
masklet generation followed by IoU/temporal filtering. The judge for
this claim returns 0.0 with the verdict ``code only shows dataset/model
placeholders\ldots not the three-phase data engine, human annotation
workflow, mask propagation correction loop, automatic masklet generation
pipeline, or timing targets.''

\textbf{Root cause.} \textit{Pipeline ordering collapse.} The data engine
is described in prose across Section~5.1 and Table~1 as a sequenced,
multi-actor protocol; the agent collapses it into a single static dataset
class, treating the engine's \emph{output} as if it were the engine
itself. This is the canonical D4 failure mode: when the temporal
relationship between phases is implicit in the prose rather than carried
by an explicit pseudocode block, the agent flattens the pipeline and the
ordering signal is lost.

\subsection{Failure Pattern Summary}

Across the four cases, a recurring pattern emerges: agents gravitate
toward \emph{familiar implementation templates} (standard ViT, temporal
concatenation, minimal evaluation scaffolding, flat dataset loaders)
rather than faithfully reproducing the paper's specific design. The
D1--D4 taxonomy pinpoints \emph{where} this drift occurs, enabling
targeted improvements.

\section{Per-Paper Semantic Alignment Scores}
\label{app:per-paper-scores}

\begin{table*}[t!]
\centering
%\footnotesize
\caption{Per-paper mean SAS across all 12 generators, sorted by overall mean.
These are paper-level means (equally weighted across papers); SAU-level means in
the main text weight each claim equally and may differ slightly.}
\label{tab:per-paper-mean}
\begin{tabular}{@{}lrrrrrr@{}}
\toprule
\textbf{Paper} & \textbf{Overall} & \textbf{Std} & \textbf{D1} & \textbf{D2} & \textbf{D3} & \textbf{D4} \\
\midrule
conformal-bayesian-quadrature   & 0.304 & 0.113 & 0.524 & 0.306 & 0.194 & 0.193 \\
mrq                             & 0.271 & 0.073 & 0.331 & 0.306 & 0.167 & 0.281 \\
olmoe                           & 0.260 & 0.096 & 0.277 & 0.296 & 0.201 & 0.264 \\
sc-fno                          & 0.257 & 0.094 & 0.319 & 0.255 & 0.196 & 0.257 \\
neural-operator-flow-matching-pde & 0.255 & 0.100 & 0.285 & 0.236 & 0.222 & 0.276 \\
ca2-vdm                         & 0.248 & 0.109 & 0.419 & 0.227 & 0.148 & 0.198 \\
voting-leaderboards             & 0.245 & 0.094 & 0.268 & 0.281 & 0.192 & 0.238 \\
robotic-world-model             & 0.243 & 0.108 & 0.352 & 0.336 & 0.160 & 0.125 \\
score                           & 0.238 & 0.066 & 0.330 & 0.285 & 0.144 & 0.193 \\
masked-diffusion-token-ordering & 0.237 & 0.104 & 0.284 & 0.230 & 0.185 & 0.250 \\
generator-augmented-flows       & 0.236 & 0.115 & 0.329 & 0.242 & 0.144 & 0.229 \\
ngpt                            & 0.236 & 0.060 & 0.299 & 0.365 & 0.122 & 0.158 \\
avg-reward-pg                   & 0.234 & 0.138 & 0.361 & 0.187 & 0.222 & 0.167 \\
hi-mar                          & 0.230 & 0.069 & 0.283 & 0.257 & 0.163 & 0.219 \\
nfig                            & 0.228 & 0.072 & 0.351 & 0.255 & 0.155 & 0.152 \\
moe-pot                         & 0.227 & 0.074 & 0.251 & 0.311 & 0.141 & 0.205 \\
luno                            & 0.225 & 0.128 & 0.306 & 0.192 & 0.205 & 0.198 \\
adjoint-matching                & 0.221 & 0.100 & 0.308 & 0.199 & 0.190 & 0.188 \\
lora-sb                         & 0.216 & 0.067 & 0.293 & 0.219 & 0.156 & 0.194 \\
gated-attention-llm             & 0.206 & 0.090 & 0.201 & 0.286 & 0.154 & 0.183 \\
emergent-planning-rl            & 0.202 & 0.084 & 0.252 & 0.211 & 0.161 & 0.184 \\
universal-neural-operators      & 0.200 & 0.060 & 0.257 & 0.190 & 0.142 & 0.212 \\
diffusion-convergence-rate      & 0.196 & 0.080 & 0.163 & 0.226 & 0.198 & 0.198 \\
wdno                            & 0.186 & 0.072 & 0.215 & 0.203 & 0.134 & 0.191 \\
ma-rlhf                         & 0.180 & 0.090 & 0.250 & 0.214 & 0.085 & 0.171 \\
navil                           & 0.179 & 0.077 & 0.210 & 0.299 & 0.098 & 0.108 \\
cara                            & 0.176 & 0.095 & 0.290 & 0.187 & 0.144 & 0.083 \\
pyramidal-flow-matching         & 0.171 & 0.065 & 0.283 & 0.153 & 0.102 & 0.146 \\
prioritized-generative-replay   & 0.158 & 0.074 & 0.212 & 0.183 & 0.148 & 0.090 \\
sam2                            & 0.158 & 0.095 & 0.193 & 0.173 & 0.124 & 0.142 \\
\midrule
\textbf{Mean (paper-level)}     & 0.221 & 0.089 & 0.290 & 0.244 & 0.160 & 0.190 \\
\bottomrule
\end{tabular}
\end{table*}

Paper difficulty is driven by specification structure, not claim volume.
The number of SAU claims correlates negatively with SAS ($r = -0.43$, $p < 0.05$,
Pearson; $\rho = -0.33$, $p = 0.08$, Spearman): papers with more claims tend
to score lower, but the relationship is noisy. For example,
\texttt{wdno} (94 claims, SAS 0.186) and \texttt{sam2} (77 claims, SAS 0.158)
diverge because \texttt{wdno}'s claims are predominantly formula-based while
\texttt{sam2}'s span heterogeneous experimental protocols. Our qualitative
inspection suggests that the easiest papers tend to concentrate specifications
in a single algorithm box and hyperparameter table, while the hardest papers
distribute specifications across sections and rely on prose over explicit
equations; however, we lack a quantitative metric for specification locality
and this observation should be treated as a hypothesis for future
investigation. D3 shows the most uniform profile across papers (IQR of
per-paper means: 0.049, lowest of the four dimensions), indicating that
experimental protocol reproduction is a near-universal difficulty independent
of domain.

Table~\ref{tab:per-paper-mean} reports per-paper mean SAS averaged across all
12 generators, with standard deviation and per-type sub-scores. The full 360-row
generator $\times$ paper matrix is available in Table~\ref{tab:full-results-matrix}.

%\clearpage

% ===================================================================
%  APPENDIX D: DISCUSSION — Pilot study, case analysis, hypotheses, findings
% ===================================================================

% ===================================================================

% \clearpage
\begingroup
\let\appendixsavedvspace\vspace
\renewcommand{\vspace}[1]{}
% === COMPREHENSIVE RESULTS TABLE (auto-generated) ===
% 360 rows = 30 papers x 12 generator configurations
% BA=BasicAgent, PC=PaperCoder, OH=OpenHands
% Timing sources: BA=batch.log, PC=artifacts/run.log, OH=artifacts/*/meta.json
% DS BA: artifacts/*/meta.json (usage.elapsed_seconds)

\section{Complete Per-Configuration Results}
\label{app:complete-results}

Table~\ref{tab:full-results-matrix} reports every paper$\times$scaffold$\times$model
SAU score (overall and D1--D4) with code generation metrics.
Times from BasicAgent: batch.log (DeepSeek BasicAgent from
artifacts/meta.json); PaperCoder: artifacts/run.log; OpenHands:
artifacts/meta.json.
The \textbf{.py} column reports the \emph{number of Python source files
generated by the agent} in the produced repository (count of \texttt{.py}
files), used as a coarse proxy for code-output volume.

\vspace{0.4cm}

{\footnotesize
\setlength{\LTpre}{0pt}
\setlength{\LTpost}{0pt}
\begin{longtable}{@{}p{1.4cm}p{1.6cm}p{2.5cm}rrrrrrr@{}}
\caption{Complete per-configuration per-paper SAU scores and code generation metrics. \textbf{Time(s)}: end-to-end agent wall-clock seconds. \textbf{.py}: number of Python source files generated by the agent in the resulting repository.}
\label{tab:full-results-matrix} \\
\toprule
\textbf{Paper} & \textbf{Scaff} & \textbf{Model} & \textbf{Overall} & \textbf{D1} & \textbf{D2} & \textbf{D3} & \textbf{D4} & \textbf{Time(s)} & \textbf{.py} \\
\midrule
\endfirsthead
\caption[]{Complete per-configuration results (continued).} \\
\toprule
\textbf{Paper} & \textbf{Scaff} & \textbf{Model} & \textbf{Overall} & \textbf{D1} & \textbf{D2} & \textbf{D3} & \textbf{D4} & \textbf{Time(s)} & \textbf{.py} \\
\midrule
\endhead
\bottomrule
\multicolumn{10}{r}{\scriptsize\textit{Continued on next page}} \\
\endfoot
\bottomrule
\endlastfoot
adjoint & BasicAgent & Claude-Sonnet-4.6 & 0.208 & 0.382 & 0.225 & 0.143 & 0.083 & 938 & 12 \\
 &  & DeepSeek-V4-Pro & 0.208 & 0.279 & 0.138 & 0.250 & 0.167 & 861 & 9 \\
 &  & Gemini-2.5-Flash & 0.136 & 0.029 & 0.287 & 0.143 & 0.083 & 678 & 3 \\
 &  & GPT-4o & 0.110 & 0.015 & 0.138 & 0.036 & 0.250 & 189 & 4 \\
 & PaperCoder & Claude-Sonnet-4.6 & 0.304 & 0.632 & 0.250 & 0.250 & 0.083 & 2408 & 12 \\
 &  & DeepSeek-V4-Pro & 0.303 & 0.471 & 0.113 & 0.214 & 0.417 & 2342 & 6 \\
 &  & Gemini-2.5-Flash & 0.332 & 0.382 & 0.325 & 0.286 & 0.333 & 1746 & 15 \\
 &  & GPT-4o & 0.199 & 0.265 & 0.150 & 0.214 & 0.167 & 1735 & 8 \\
 & OpenHands & Claude-Sonnet-4.6 & 0.262 & 0.382 & 0.250 & 0.250 & 0.167 & 1142 & 12 \\
 &  & DeepSeek-V4-Pro & 0.330 & 0.485 & 0.300 & 0.286 & 0.250 & 1345 & 16 \\
 &  & Gemini-2.5-Flash & 0.261 & 0.368 & 0.212 & 0.214 & 0.250 & 433 & 6 \\
 &  & GPT-4o & 0.000 & 0.000 & 0.000 & 0.000 & 0.000 & 30 & 0 \\
\addlinespace
avg-rwd & BasicAgent & Claude-Sonnet-4.6 & 0.321 & 0.667 & 0.286 & 0.250 & 0.083 & 1305 & 5 \\
 &  & DeepSeek-V4-Pro & 0.226 & 0.417 & 0.321 & 0.083 & 0.083 & 934 & 11 \\
 &  & Gemini-2.5-Flash & 0.155 & 0.000 & 0.286 & 0.000 & 0.333 & 580 & 6 \\
 &  & GPT-4o & 0.000 & 0.000 & 0.000 & 0.000 & 0.000 & 132 & 2 \\
 & PaperCoder & Claude-Sonnet-4.6 & 0.455 & 0.417 & 0.321 & 0.667 & 0.417 & 1407 & 9 \\
 &  & DeepSeek-V4-Pro & 0.303 & 0.556 & 0.071 & 0.500 & 0.083 & 1460 & 5 \\
 &  & Gemini-2.5-Flash & 0.150 & 0.278 & 0.071 & 0.250 & 0.000 & 454 & 7 \\
 &  & GPT-4o & 0.192 & 0.278 & 0.071 & 0.167 & 0.250 & 758 & 7 \\
 & OpenHands & Claude-Sonnet-4.6 & 0.262 & 0.417 & 0.214 & 0.167 & 0.250 & 798 & 9 \\
 &  & DeepSeek-V4-Pro & 0.402 & 0.667 & 0.357 & 0.333 & 0.250 & 782 & 7 \\
 &  & Gemini-2.5-Flash & 0.318 & 0.639 & 0.214 & 0.250 & 0.167 & 296 & 4 \\
 &  & GPT-4o & 0.030 & 0.000 & 0.036 & 0.000 & 0.083 & 407 & 4 \\
\addlinespace
ca2-vdm & BasicAgent & Claude-Sonnet-4.6 & 0.302 & 0.614 & 0.250 & 0.094 & 0.250 & 1234 & 20 \\
 &  & DeepSeek-V4-Pro & 0.367 & 0.614 & 0.325 & 0.250 & 0.281 & 1556 & 11 \\
 &  & Gemini-2.5-Flash & 0.059 & 0.023 & 0.150 & 0.031 & 0.031 & 2814 & 1 \\
 &  & GPT-4o & 0.025 & 0.000 & 0.100 & 0.000 & 0.000 & 167 & 5 \\
 & PaperCoder & Claude-Sonnet-4.6 & 0.362 & 0.750 & 0.325 & 0.188 & 0.188 & 1992 & 9 \\
 &  & DeepSeek-V4-Pro & 0.244 & 0.500 & 0.225 & 0.094 & 0.156 & 7168 & 15 \\
 &  & Gemini-2.5-Flash & 0.290 & 0.477 & 0.275 & 0.156 & 0.250 & 1386 & 11 \\
 &  & GPT-4o & 0.255 & 0.352 & 0.200 & 0.219 & 0.250 & 931 & 8 \\
 & OpenHands & Claude-Sonnet-4.6 & 0.331 & 0.523 & 0.300 & 0.250 & 0.250 & 1377 & 9 \\
 &  & DeepSeek-V4-Pro & 0.298 & 0.443 & 0.250 & 0.219 & 0.281 & 1689 & 9 \\
 &  & Gemini-2.5-Flash & 0.257 & 0.420 & 0.200 & 0.188 & 0.219 & 704 & 6 \\
 &  & GPT-4o & 0.186 & 0.307 & 0.125 & 0.094 & 0.219 & 348 & 5 \\
\addlinespace
cara & BasicAgent & Claude-Sonnet-4.6 & 0.205 & 0.365 & 0.315 & 0.056 & 0.083 & 1169 & 29 \\
 &  & DeepSeek-V4-Pro & 0.241 & 0.567 & 0.232 & 0.083 & 0.083 & 1308 & 20 \\
 &  & Gemini-2.5-Flash & 0.117 & 0.144 & 0.130 & 0.111 & 0.083 & 1379 & 13 \\
 &  & GPT-4o & 0.012 & 0.038 & 0.009 & 0.000 & 0.000 & 94 & 4 \\
 & PaperCoder & Claude-Sonnet-4.6 & 0.292 & 0.558 & 0.361 & 0.250 & 0.000 & 4347 & 20 \\
 &  & DeepSeek-V4-Pro & 0.158 & 0.288 & 0.148 & 0.111 & 0.083 & 3005 & 8 \\
 &  & Gemini-2.5-Flash & 0.223 & 0.279 & 0.194 & 0.250 & 0.167 & 1421 & 16 \\
 &  & GPT-4o & 0.171 & 0.240 & 0.139 & 0.222 & 0.083 & 843 & 7 \\
 & OpenHands & Claude-Sonnet-4.6 & 0.275 & 0.423 & 0.259 & 0.250 & 0.167 & 1450 & 19 \\
 &  & DeepSeek-V4-Pro & 0.254 & 0.375 & 0.250 & 0.222 & 0.167 & 1359 & 10 \\
 &  & Gemini-2.5-Flash & 0.164 & 0.202 & 0.204 & 0.167 & 0.083 & 551 & 6 \\
 &  & GPT-4o & 0.000 & 0.000 & 0.000 & 0.000 & 0.000 & 200 & 0 \\
\addlinespace
conf-bq & BasicAgent & Claude-Sonnet-4.6 & 0.374 & 0.870 & 0.417 & 0.083 & 0.125 & 950 & 8 \\
 &  & DeepSeek-V4-Pro & 0.430 & 0.700 & 0.583 & 0.250 & 0.188 & 840 & 10 \\
 &  & Gemini-2.5-Flash & 0.244 & 0.330 & 0.292 & 0.167 & 0.188 & 527 & 5 \\
 &  & GPT-4o & 0.000 & 0.000 & 0.000 & 0.000 & 0.000 & 121 & 3 \\
 & PaperCoder & Claude-Sonnet-4.6 & 0.425 & 0.930 & 0.292 & 0.167 & 0.312 & 2019 & 12 \\
 &  & DeepSeek-V4-Pro & 0.353 & 0.580 & 0.417 & 0.167 & 0.250 & 1926 & 7 \\
 &  & Gemini-2.5-Flash & 0.330 & 0.550 & 0.333 & 0.250 & 0.188 & 873 & 8 \\
 &  & GPT-4o & 0.283 & 0.340 & 0.292 & 0.250 & 0.250 & 754 & 7 \\
 & OpenHands & Claude-Sonnet-4.6 & 0.332 & 0.640 & 0.250 & 0.250 & 0.188 & 1323 & 10 \\
 &  & DeepSeek-V4-Pro & 0.305 & 0.490 & 0.208 & 0.333 & 0.188 & 729 & 7 \\
 &  & Gemini-2.5-Flash & 0.336 & 0.490 & 0.417 & 0.250 & 0.188 & 437 & 9 \\
 &  & GPT-4o & 0.238 & 0.370 & 0.167 & 0.167 & 0.250 & 262 & 3 \\
\addlinespace
diff-conv & BasicAgent & Claude-Sonnet-4.6 & 0.237 & 0.239 & 0.333 & 0.125 & 0.250 & 1025 & 10 \\
 &  & DeepSeek-V4-Pro & 0.233 & 0.159 & 0.333 & 0.188 & 0.250 & 873 & 10 \\
 &  & Gemini-2.5-Flash & 0.221 & 0.114 & 0.271 & 0.250 & 0.250 & 324 & 2 \\
 &  & GPT-4o & 0.088 & 0.000 & 0.167 & 0.000 & 0.188 & 112 & 3 \\
 & PaperCoder & Claude-Sonnet-4.6 & 0.263 & 0.239 & 0.188 & 0.500 & 0.125 & 1558 & 9 \\
 &  & DeepSeek-V4-Pro & 0.280 & 0.182 & 0.312 & 0.312 & 0.312 & 1800 & 6 \\
 &  & Gemini-2.5-Flash & 0.246 & 0.193 & 0.292 & 0.188 & 0.312 & 579 & 6 \\
 &  & GPT-4o & 0.170 & 0.136 & 0.229 & 0.125 & 0.188 & 1230 & 8 \\
 & OpenHands & Claude-Sonnet-4.6 & 0.236 & 0.341 & 0.229 & 0.188 & 0.188 & 1966 & 12 \\
 &  & DeepSeek-V4-Pro & 0.202 & 0.182 & 0.188 & 0.250 & 0.188 & 1703 & 7 \\
 &  & Gemini-2.5-Flash & 0.178 & 0.171 & 0.167 & 0.250 & 0.125 & 366 & 6 \\
 &  & GPT-4o & 0.000 & 0.000 & 0.000 & 0.000 & 0.000 & 545 & 0 \\
\addlinespace
emergent & BasicAgent & Claude-Sonnet-4.6 & 0.146 & 0.225 & 0.118 & 0.073 & 0.167 & 1204 & 23 \\
 &  & DeepSeek-V4-Pro & 0.237 & 0.325 & 0.294 & 0.162 & 0.167 & 1722 & 12 \\
 &  & Gemini-2.5-Flash & 0.041 & 0.037 & 0.044 & 0.000 & 0.083 & 2940 & 1 \\
 &  & GPT-4o & 0.113 & 0.062 & 0.088 & 0.176 & 0.125 & 100 & 3 \\
 & PaperCoder & Claude-Sonnet-4.6 & 0.261 & 0.525 & 0.176 & 0.132 & 0.208 & 1400 & 8 \\
 &  & DeepSeek-V4-Pro & 0.265 & 0.312 & 0.294 & 0.162 & 0.292 & 4746 & 10 \\
 &  & Gemini-2.5-Flash & 0.269 & 0.312 & 0.279 & 0.235 & 0.250 & 2631 & 18 \\
 &  & GPT-4o & 0.200 & 0.150 & 0.235 & 0.206 & 0.208 & 1314 & 9 \\
 & OpenHands & Claude-Sonnet-4.6 & 0.280 & 0.312 & 0.309 & 0.250 & 0.250 & 1214 & 28 \\
 &  & DeepSeek-V4-Pro & 0.263 & 0.312 & 0.324 & 0.206 & 0.208 & 1664 & 24 \\
 &  & Gemini-2.5-Flash & 0.266 & 0.325 & 0.294 & 0.235 & 0.208 & 932 & 8 \\
 &  & GPT-4o & 0.082 & 0.125 & 0.073 & 0.088 & 0.042 & 548 & 4 \\
\addlinespace
gated-attn & BasicAgent & Claude-Sonnet-4.6 & 0.226 & 0.233 & 0.237 & 0.132 & 0.300 & 834 & 13 \\
 &  & DeepSeek-V4-Pro & 0.194 & 0.283 & 0.263 & 0.103 & 0.125 & 1725 & 19 \\
 &  & Gemini-2.5-Flash & 0.087 & 0.050 & 0.250 & 0.000 & 0.050 & 189 & 1 \\
 &  & GPT-4o & 0.100 & 0.033 & 0.197 & 0.044 & 0.125 & 109 & 3 \\
 & PaperCoder & Claude-Sonnet-4.6 & 0.408 & 0.400 & 0.645 & 0.338 & 0.250 & 2969 & 12 \\
 &  & DeepSeek-V4-Pro & 0.189 & 0.167 & 0.276 & 0.162 & 0.150 & 3107 & 8 \\
 &  & Gemini-2.5-Flash & 0.207 & 0.200 & 0.276 & 0.176 & 0.175 & 1235 & 11 \\
 &  & GPT-4o & 0.161 & 0.167 & 0.197 & 0.103 & 0.175 & 640 & 6 \\
 & OpenHands & Claude-Sonnet-4.6 & 0.284 & 0.333 & 0.342 & 0.235 & 0.225 & 879 & 8 \\
 &  & DeepSeek-V4-Pro & 0.260 & 0.267 & 0.342 & 0.206 & 0.225 & 1328 & 9 \\
 &  & Gemini-2.5-Flash & 0.247 & 0.250 & 0.263 & 0.250 & 0.225 & 681 & 6 \\
 &  & GPT-4o & 0.114 & 0.033 & 0.145 & 0.103 & 0.175 & 279 & 6 \\
\addlinespace
gen-flows & BasicAgent & Claude-Sonnet-4.6 & 0.432 & 0.518 & 0.356 & 0.229 & 0.625 & 1088 & 15 \\
 &  & DeepSeek-V4-Pro & 0.255 & 0.500 & 0.394 & 0.125 & 0.000 & 1711 & 13 \\
 &  & Gemini-2.5-Flash & 0.184 & 0.107 & 0.192 & 0.062 & 0.375 & 1562 & 5 \\
 &  & GPT-4o & 0.095 & 0.089 & 0.038 & 0.125 & 0.125 & 162 & 4 \\
 & PaperCoder & Claude-Sonnet-4.6 & 0.286 & 0.446 & 0.260 & 0.188 & 0.250 & 2358 & 11 \\
 &  & DeepSeek-V4-Pro & 0.145 & 0.179 & 0.192 & 0.083 & 0.125 & 2859 & 8 \\
 &  & Gemini-2.5-Flash & 0.260 & 0.375 & 0.269 & 0.146 & 0.250 & 703 & 7 \\
 &  & GPT-4o & 0.193 & 0.196 & 0.202 & 0.125 & 0.250 & 724 & 6 \\
 & OpenHands & Claude-Sonnet-4.6 & 0.319 & 0.500 & 0.298 & 0.229 & 0.250 & 1120 & 11 \\
 &  & DeepSeek-V4-Pro & 0.363 & 0.482 & 0.365 & 0.229 & 0.375 & 956 & 10 \\
 &  & Gemini-2.5-Flash & 0.279 & 0.518 & 0.308 & 0.167 & 0.125 & 254 & 6 \\
 &  & GPT-4o & 0.021 & 0.036 & 0.029 & 0.021 & 0.000 & 379 & 7 \\
\addlinespace
hi-mar & BasicAgent & Claude-Sonnet-4.6 & 0.304 & 0.384 & 0.375 & 0.208 & 0.250 & 956 & 13 \\
 &  & DeepSeek-V4-Pro & 0.189 & 0.214 & 0.167 & 0.125 & 0.250 & 1451 & 12 \\
 &  & Gemini-2.5-Flash & 0.165 & 0.161 & 0.229 & 0.083 & 0.188 & 1010 & 3 \\
 &  & GPT-4o & 0.066 & 0.054 & 0.167 & 0.042 & 0.000 & 109 & 4 \\
 & PaperCoder & Claude-Sonnet-4.6 & 0.293 & 0.527 & 0.250 & 0.208 & 0.188 & 3148 & 14 \\
 &  & DeepSeek-V4-Pro & 0.238 & 0.286 & 0.333 & 0.083 & 0.250 & 4923 & 11 \\
 &  & Gemini-2.5-Flash & 0.266 & 0.312 & 0.292 & 0.208 & 0.250 & 1011 & 9 \\
 &  & GPT-4o & 0.258 & 0.304 & 0.229 & 0.250 & 0.250 & 635 & 6 \\
 & OpenHands & Claude-Sonnet-4.6 & 0.287 & 0.420 & 0.271 & 0.208 & 0.250 & 788 & 7 \\
 &  & DeepSeek-V4-Pro & 0.261 & 0.232 & 0.354 & 0.208 & 0.250 & 1913 & 14 \\
 &  & Gemini-2.5-Flash & 0.257 & 0.384 & 0.229 & 0.167 & 0.250 & 1743 & 6 \\
 &  & GPT-4o & 0.180 & 0.116 & 0.188 & 0.167 & 0.250 & 408 & 6 \\
\addlinespace
lora-sb & BasicAgent & Claude-Sonnet-4.6 & 0.310 & 0.354 & 0.304 & 0.250 & 0.333 & 776 & 8 \\
 &  & DeepSeek-V4-Pro & 0.207 & 0.271 & 0.286 & 0.188 & 0.083 & 1135 & 8 \\
 &  & Gemini-2.5-Flash & 0.145 & 0.188 & 0.214 & 0.094 & 0.083 & 397 & 4 \\
 &  & GPT-4o & 0.119 & 0.125 & 0.143 & 0.125 & 0.083 & 120 & 2 \\
 & PaperCoder & Claude-Sonnet-4.6 & 0.214 & 0.458 & 0.054 & 0.094 & 0.250 & 1900 & 11 \\
 &  & DeepSeek-V4-Pro & 0.224 & 0.333 & 0.304 & 0.094 & 0.167 & 2478 & 7 \\
 &  & Gemini-2.5-Flash & 0.249 & 0.312 & 0.214 & 0.219 & 0.250 & 604 & 6 \\
 &  & GPT-4o & 0.202 & 0.271 & 0.161 & 0.125 & 0.250 & 908 & 6 \\
 & OpenHands & Claude-Sonnet-4.6 & 0.262 & 0.333 & 0.214 & 0.250 & 0.250 & 890 & 12 \\
 &  & DeepSeek-V4-Pro & 0.278 & 0.333 & 0.339 & 0.188 & 0.250 & 1224 & 7 \\
 &  & Gemini-2.5-Flash & 0.284 & 0.500 & 0.250 & 0.219 & 0.167 & 221 & 6 \\
 &  & GPT-4o & 0.096 & 0.042 & 0.143 & 0.031 & 0.167 & 335 & 6 \\
\addlinespace
luno & BasicAgent & Claude-Sonnet-4.6 & 0.279 & 0.450 & 0.250 & 0.167 & 0.250 & 990 & 15 \\
 &  & DeepSeek-V4-Pro & 0.288 & 0.350 & 0.260 & 0.292 & 0.250 & 1231 & 16 \\
 &  & Gemini-2.5-Flash & 0.062 & 0.050 & 0.156 & 0.000 & 0.042 & 648 & 2 \\
 &  & GPT-4o & 0.026 & 0.000 & 0.104 & 0.000 & 0.000 & 187 & 5 \\
 & PaperCoder & Claude-Sonnet-4.6 & 0.433 & 0.525 & 0.208 & 0.583 & 0.417 & 3295 & 17 \\
 &  & DeepSeek-V4-Pro & 0.254 & 0.400 & 0.198 & 0.167 & 0.250 & 4540 & 9 \\
 &  & Gemini-2.5-Flash & 0.276 & 0.375 & 0.188 & 0.292 & 0.250 & 1462 & 9 \\
 &  & GPT-4o & 0.255 & 0.375 & 0.229 & 0.208 & 0.208 & 728 & 6 \\
 & OpenHands & Claude-Sonnet-4.6 & 0.248 & 0.275 & 0.219 & 0.250 & 0.250 & 1545 & 21 \\
 &  & DeepSeek-V4-Pro & 0.306 & 0.400 & 0.281 & 0.292 & 0.250 & 978 & 9 \\
 &  & Gemini-2.5-Flash & 0.275 & 0.475 & 0.208 & 0.208 & 0.208 & 845 & 8 \\
 &  & GPT-4o & 0.000 & 0.000 & 0.000 & 0.000 & 0.000 & 413 & 0 \\
\addlinespace
ma-rlhf & BasicAgent & Claude-Sonnet-4.6 & 0.220 & 0.306 & 0.355 & 0.068 & 0.150 & 874 & 11 \\
 &  & DeepSeek-V4-Pro & 0.269 & 0.472 & 0.197 & 0.204 & 0.200 & 1206 & 11 \\
 &  & Gemini-2.5-Flash & 0.131 & 0.083 & 0.171 & 0.068 & 0.200 & 845 & 5 \\
 &  & GPT-4o & 0.013 & 0.028 & 0.000 & 0.023 & 0.000 & 214 & 5 \\
 & PaperCoder & Claude-Sonnet-4.6 & 0.238 & 0.444 & 0.210 & 0.045 & 0.250 & 2571 & 11 \\
 &  & DeepSeek-V4-Pro & 0.177 & 0.222 & 0.342 & 0.045 & 0.100 & 3128 & 7 \\
 &  & Gemini-2.5-Flash & 0.236 & 0.361 & 0.263 & 0.068 & 0.250 & 1548 & 11 \\
 &  & GPT-4o & 0.171 & 0.139 & 0.276 & 0.068 & 0.200 & 619 & 7 \\
 & OpenHands & Claude-Sonnet-4.6 & 0.263 & 0.417 & 0.250 & 0.136 & 0.250 & 862 & 10 \\
 &  & DeepSeek-V4-Pro & 0.242 & 0.250 & 0.263 & 0.204 & 0.250 & 1112 & 9 \\
 &  & Gemini-2.5-Flash & 0.201 & 0.278 & 0.237 & 0.091 & 0.200 & 307 & 6 \\
 &  & GPT-4o & 0.000 & 0.000 & 0.000 & 0.000 & 0.000 & 250 & 0 \\
\addlinespace
masked-diff & BasicAgent & Claude-Sonnet-4.6 & 0.280 & 0.440 & 0.271 & 0.222 & 0.188 & 1027 & 13 \\
 &  & DeepSeek-V4-Pro & 0.369 & 0.405 & 0.458 & 0.361 & 0.250 & 1228 & 22 \\
 &  & Gemini-2.5-Flash & 0.179 & 0.052 & 0.156 & 0.194 & 0.312 & 752 & 7 \\
 &  & GPT-4o & 0.127 & 0.009 & 0.146 & 0.167 & 0.188 & 137 & 3 \\
 & PaperCoder & Claude-Sonnet-4.6 & 0.398 & 0.526 & 0.365 & 0.139 & 0.562 & 3336 & 15 \\
 &  & DeepSeek-V4-Pro & 0.282 & 0.414 & 0.292 & 0.111 & 0.312 & 2797 & 8 \\
 &  & Gemini-2.5-Flash & 0.242 & 0.336 & 0.188 & 0.194 & 0.250 & 1229 & 14 \\
 &  & GPT-4o & 0.192 & 0.198 & 0.188 & 0.194 & 0.188 & 756 & 7 \\
 & OpenHands & Claude-Sonnet-4.6 & 0.283 & 0.440 & 0.219 & 0.222 & 0.250 & 1564 & 17 \\
 &  & DeepSeek-V4-Pro & 0.268 & 0.353 & 0.219 & 0.250 & 0.250 & 1089 & 8 \\
 &  & Gemini-2.5-Flash & 0.217 & 0.233 & 0.219 & 0.167 & 0.250 & 252 & 7 \\
 &  & GPT-4o & 0.010 & 0.000 & 0.042 & 0.000 & 0.000 & 390 & 6 \\
\addlinespace
moe-pot & BasicAgent & Claude-Sonnet-4.6 & 0.254 & 0.273 & 0.383 & 0.147 & 0.214 & 734 & 11 \\
 &  & DeepSeek-V4-Pro & 0.244 & 0.239 & 0.350 & 0.103 & 0.286 & 1065 & 10 \\
 &  & Gemini-2.5-Flash & 0.106 & 0.057 & 0.200 & 0.059 & 0.107 & 492 & 4 \\
 &  & GPT-4o & 0.105 & 0.091 & 0.183 & 0.073 & 0.071 & 192 & 3 \\
 & PaperCoder & Claude-Sonnet-4.6 & 0.299 & 0.466 & 0.333 & 0.147 & 0.250 & 3643 & 19 \\
 &  & DeepSeek-V4-Pro & 0.233 & 0.284 & 0.417 & 0.088 & 0.143 & 2990 & 7 \\
 &  & Gemini-2.5-Flash & 0.319 & 0.455 & 0.317 & 0.221 & 0.286 & 1239 & 10 \\
 &  & GPT-4o & 0.242 & 0.239 & 0.267 & 0.176 & 0.286 & 946 & 6 \\
 & OpenHands & Claude-Sonnet-4.6 & 0.281 & 0.250 & 0.467 & 0.191 & 0.214 & 930 & 8 \\
 &  & DeepSeek-V4-Pro & 0.287 & 0.295 & 0.333 & 0.235 & 0.286 & 1350 & 6 \\
 &  & Gemini-2.5-Flash & 0.227 & 0.216 & 0.350 & 0.162 & 0.179 & 427 & 12 \\
 &  & GPT-4o & 0.128 & 0.148 & 0.133 & 0.088 & 0.143 & 372 & 6 \\
\addlinespace
mrq & BasicAgent & Claude-Sonnet-4.6 & 0.274 & 0.306 & 0.279 & 0.200 & 0.312 & 1384 & 12 \\
 &  & DeepSeek-V4-Pro & 0.280 & 0.333 & 0.338 & 0.200 & 0.250 & 770 & 10 \\
 &  & Gemini-2.5-Flash & 0.305 & 0.389 & 0.368 & 0.150 & 0.312 & 1938 & 7 \\
 &  & GPT-4o & 0.108 & 0.083 & 0.162 & 0.000 & 0.188 & 97 & 3 \\
 & PaperCoder & Claude-Sonnet-4.6 & 0.320 & 0.417 & 0.412 & 0.200 & 0.250 & 2051 & 7 \\
 &  & DeepSeek-V4-Pro & 0.279 & 0.333 & 0.221 & 0.250 & 0.312 & 3639 & 9 \\
 &  & Gemini-2.5-Flash & 0.328 & 0.417 & 0.382 & 0.200 & 0.312 & 1341 & 9 \\
 &  & GPT-4o & 0.280 & 0.361 & 0.309 & 0.200 & 0.250 & 1490 & 7 \\
 & OpenHands & Claude-Sonnet-4.6 & 0.290 & 0.417 & 0.294 & 0.200 & 0.250 & 2830 & 6 \\
 &  & DeepSeek-V4-Pro & 0.335 & 0.333 & 0.368 & 0.200 & 0.438 & 2275 & 8 \\
 &  & Gemini-2.5-Flash & 0.320 & 0.389 & 0.368 & 0.150 & 0.375 & 615 & 6 \\
 &  & GPT-4o & 0.137 & 0.194 & 0.176 & 0.050 & 0.125 & 1711 & 5 \\
\addlinespace
navil & BasicAgent & Claude-Sonnet-4.6 & 0.290 & 0.455 & 0.479 & 0.100 & 0.125 & 2281 & 14 \\
 &  & DeepSeek-V4-Pro & 0.204 & 0.136 & 0.479 & 0.050 & 0.150 & 1185 & 12 \\
 &  & Gemini-2.5-Flash & 0.117 & 0.159 & 0.208 & 0.025 & 0.075 & 628 & 5 \\
 &  & GPT-4o & 0.039 & 0.068 & 0.062 & 0.025 & 0.000 & 80 & 2 \\
 & PaperCoder & Claude-Sonnet-4.6 & 0.245 & 0.318 & 0.438 & 0.025 & 0.200 & 3471 & 17 \\
 &  & DeepSeek-V4-Pro & 0.182 & 0.159 & 0.396 & 0.050 & 0.125 & 4998 & 13 \\
 &  & Gemini-2.5-Flash & 0.224 & 0.295 & 0.250 & 0.200 & 0.150 & 1742 & 11 \\
 &  & GPT-4o & 0.103 & 0.091 & 0.146 & 0.125 & 0.050 & 1094 & 7 \\
 & OpenHands & Claude-Sonnet-4.6 & 0.229 & 0.295 & 0.271 & 0.225 & 0.125 & 836 & 7 \\
 &  & DeepSeek-V4-Pro & 0.246 & 0.250 & 0.333 & 0.200 & 0.200 & 1351 & 8 \\
 &  & Gemini-2.5-Flash & 0.185 & 0.227 & 0.312 & 0.125 & 0.075 & 293 & 6 \\
 &  & GPT-4o & 0.082 & 0.068 & 0.208 & 0.025 & 0.025 & 500 & 4 \\
\addlinespace
neural-op & BasicAgent & Claude-Sonnet-4.6 & 0.432 & 0.481 & 0.433 & 0.250 & 0.562 & 1223 & 14 \\
 &  & DeepSeek-V4-Pro & 0.281 & 0.308 & 0.317 & 0.250 & 0.250 & 1493 & 19 \\
 &  & Gemini-2.5-Flash & 0.119 & 0.077 & 0.150 & 0.125 & 0.125 & 2846 & 3 \\
 &  & GPT-4o & 0.133 & 0.096 & 0.083 & 0.167 & 0.188 & 170 & 5 \\
 & PaperCoder & Claude-Sonnet-4.6 & 0.403 & 0.577 & 0.200 & 0.333 & 0.500 & 3553 & 16 \\
 &  & DeepSeek-V4-Pro & 0.293 & 0.423 & 0.250 & 0.250 & 0.250 & 3915 & 10 \\
 &  & Gemini-2.5-Flash & 0.294 & 0.308 & 0.367 & 0.250 & 0.250 & 2194 & 12 \\
 &  & GPT-4o & 0.172 & 0.154 & 0.117 & 0.167 & 0.250 & 1087 & 7 \\
 & OpenHands & Claude-Sonnet-4.6 & 0.256 & 0.288 & 0.233 & 0.250 & 0.250 & 1932 & 9 \\
 &  & DeepSeek-V4-Pro & 0.290 & 0.327 & 0.333 & 0.250 & 0.250 & 2812 & 5 \\
 &  & Gemini-2.5-Flash & 0.240 & 0.288 & 0.233 & 0.250 & 0.188 & 319 & 6 \\
 &  & GPT-4o & 0.147 & 0.096 & 0.117 & 0.125 & 0.250 & 379 & 7 \\
\addlinespace
nfig & BasicAgent & Claude-Sonnet-4.6 & 0.252 & 0.464 & 0.222 & 0.143 & 0.179 & 1195 & 12 \\
 &  & DeepSeek-V4-Pro & 0.219 & 0.500 & 0.056 & 0.107 & 0.214 & 1478 & 14 \\
 &  & Gemini-2.5-Flash & 0.156 & 0.179 & 0.194 & 0.143 & 0.107 & 1251 & 3 \\
 &  & GPT-4o & 0.102 & 0.036 & 0.194 & 0.071 & 0.107 & 96 & 8 \\
 & PaperCoder & Claude-Sonnet-4.6 & 0.220 & 0.500 & 0.167 & 0.071 & 0.143 & 3196 & 19 \\
 &  & DeepSeek-V4-Pro & 0.285 & 0.536 & 0.389 & 0.071 & 0.143 & 4051 & 10 \\
 &  & Gemini-2.5-Flash & 0.283 & 0.357 & 0.417 & 0.214 & 0.143 & 1774 & 10 \\
 &  & GPT-4o & 0.203 & 0.214 & 0.278 & 0.214 & 0.107 & 1042 & 7 \\
 & OpenHands & Claude-Sonnet-4.6 & 0.279 & 0.429 & 0.222 & 0.250 & 0.214 & 1052 & 15 \\
 &  & DeepSeek-V4-Pro & 0.295 & 0.464 & 0.250 & 0.250 & 0.214 & 1944 & 15 \\
 &  & Gemini-2.5-Flash & 0.325 & 0.429 & 0.444 & 0.250 & 0.179 & 395 & 6 \\
 &  & GPT-4o & 0.118 & 0.107 & 0.222 & 0.071 & 0.071 & 348 & 4 \\
\addlinespace
ngpt & BasicAgent & Claude-Sonnet-4.6 & 0.260 & 0.333 & 0.420 & 0.107 & 0.179 & 1776 & 6 \\
 &  & DeepSeek-V4-Pro & 0.287 & 0.441 & 0.420 & 0.107 & 0.179 & 1085 & 7 \\
 &  & Gemini-2.5-Flash & 0.248 & 0.238 & 0.470 & 0.107 & 0.179 & 231 & 6 \\
 &  & GPT-4o & 0.130 & 0.059 & 0.280 & 0.071 & 0.107 & 171 & 7 \\
 & PaperCoder & Claude-Sonnet-4.6 & 0.271 & 0.381 & 0.380 & 0.143 & 0.179 & 1953 & 8 \\
 &  & DeepSeek-V4-Pro & 0.215 & 0.357 & 0.360 & 0.071 & 0.071 & 3504 & 8 \\
 &  & Gemini-2.5-Flash & 0.305 & 0.369 & 0.460 & 0.214 & 0.179 & 1218 & 7 \\
 &  & GPT-4o & 0.186 & 0.214 & 0.280 & 0.143 & 0.107 & 998 & 6 \\
 & OpenHands & Claude-Sonnet-4.6 & 0.291 & 0.381 & 0.390 & 0.179 & 0.214 & 826 & 7 \\
 &  & DeepSeek-V4-Pro & 0.284 & 0.417 & 0.400 & 0.107 & 0.214 & 1066 & 6 \\
 &  & Gemini-2.5-Flash & 0.222 & 0.309 & 0.330 & 0.107 & 0.143 & 159 & 7 \\
 &  & GPT-4o & 0.131 & 0.083 & 0.190 & 0.107 & 0.143 & 322 & 6 \\
\addlinespace
olmoe & BasicAgent & Claude-Sonnet-4.6 & 0.375 & 0.388 & 0.362 & 0.250 & 0.500 & 1215 & 11 \\
 &  & DeepSeek-V4-Pro & 0.412 & 0.338 & 0.517 & 0.208 & 0.583 & 1100 & 16 \\
 &  & Gemini-2.5-Flash & 0.130 & 0.075 & 0.112 & 0.167 & 0.167 & 407 & 4 \\
 &  & GPT-4o & 0.126 & 0.125 & 0.129 & 0.083 & 0.167 & 139 & 2 \\
 & PaperCoder & Claude-Sonnet-4.6 & 0.260 & 0.375 & 0.500 & 0.167 & 0.000 & 4993 & 24 \\
 &  & DeepSeek-V4-Pro & 0.336 & 0.388 & 0.414 & 0.208 & 0.333 & 5013 & 14 \\
 &  & Gemini-2.5-Flash & 0.289 & 0.312 & 0.259 & 0.250 & 0.333 & 1392 & 16 \\
 &  & GPT-4o & 0.239 & 0.225 & 0.233 & 0.250 & 0.250 & 949 & 7 \\
 & OpenHands & Claude-Sonnet-4.6 & 0.258 & 0.275 & 0.259 & 0.250 & 0.250 & 756 & 9 \\
 &  & DeepSeek-V4-Pro & 0.289 & 0.338 & 0.319 & 0.250 & 0.250 & 1566 & 23 \\
 &  & Gemini-2.5-Flash & 0.287 & 0.362 & 0.328 & 0.208 & 0.250 & 806 & 6 \\
 &  & GPT-4o & 0.114 & 0.125 & 0.121 & 0.125 & 0.083 & 645 & 6 \\
\addlinespace
pgr & BasicAgent & Claude-Sonnet-4.6 & 0.095 & 0.120 & 0.089 & 0.088 & 0.083 & 876 & 10 \\
 &  & DeepSeek-V4-Pro & 0.224 & 0.326 & 0.250 & 0.235 & 0.083 & 1375 & 17 \\
 &  & Gemini-2.5-Flash & 0.136 & 0.163 & 0.179 & 0.118 & 0.083 & 2333 & 12 \\
 &  & GPT-4o & 0.123 & 0.022 & 0.214 & 0.088 & 0.167 & 90 & 4 \\
 & PaperCoder & Claude-Sonnet-4.6 & 0.233 & 0.435 & 0.321 & 0.176 & 0.000 & 3958 & 18 \\
 &  & DeepSeek-V4-Pro & 0.093 & 0.152 & 0.161 & 0.059 & 0.000 & 3158 & 9 \\
 &  & Gemini-2.5-Flash & 0.212 & 0.293 & 0.196 & 0.191 & 0.167 & 1879 & 18 \\
 &  & GPT-4o & 0.134 & 0.141 & 0.179 & 0.132 & 0.083 & 1017 & 10 \\
 & OpenHands & Claude-Sonnet-4.6 & 0.209 & 0.337 & 0.196 & 0.221 & 0.083 & 1140 & 12 \\
 &  & DeepSeek-V4-Pro & 0.242 & 0.304 & 0.232 & 0.265 & 0.167 & 1867 & 12 \\
 &  & Gemini-2.5-Flash & 0.200 & 0.250 & 0.179 & 0.206 & 0.167 & 454 & 5 \\
 &  & GPT-4o & 0.000 & 0.000 & 0.000 & 0.000 & 0.000 & 184 & 0 \\
\addlinespace
pyramidal & BasicAgent & Claude-Sonnet-4.6 & 0.230 & 0.461 & 0.162 & 0.139 & 0.159 & 1027 & 17 \\
 &  & DeepSeek-V4-Pro & 0.205 & 0.430 & 0.147 & 0.083 & 0.159 & 1710 & 20 \\
 &  & Gemini-2.5-Flash & 0.070 & 0.039 & 0.118 & 0.056 & 0.068 & 2921 & 3 \\
 &  & GPT-4o & 0.074 & 0.039 & 0.118 & 0.028 & 0.114 & 114 & 4 \\
 & PaperCoder & Claude-Sonnet-4.6 & 0.243 & 0.500 & 0.235 & 0.167 & 0.068 & 4097 & 20 \\
 &  & DeepSeek-V4-Pro & 0.189 & 0.281 & 0.235 & 0.056 & 0.182 & 4950 & 9 \\
 &  & Gemini-2.5-Flash & 0.179 & 0.328 & 0.162 & 0.111 & 0.114 & 1512 & 10 \\
 &  & GPT-4o & 0.156 & 0.250 & 0.132 & 0.083 & 0.159 & 948 & 8 \\
 & OpenHands & Claude-Sonnet-4.6 & 0.227 & 0.375 & 0.132 & 0.194 & 0.204 & 1280 & 22 \\
 &  & DeepSeek-V4-Pro & 0.231 & 0.352 & 0.176 & 0.167 & 0.227 & 1682 & 11 \\
 &  & Gemini-2.5-Flash & 0.175 & 0.312 & 0.118 & 0.111 & 0.159 & 282 & 6 \\
 &  & GPT-4o & 0.075 & 0.031 & 0.103 & 0.028 & 0.136 & 428 & 7 \\
\addlinespace
robotic-wm & BasicAgent & Claude-Sonnet-4.6 & 0.310 & 0.463 & 0.571 & 0.125 & 0.083 & 1163 & 17 \\
 &  & DeepSeek-V4-Pro & 0.300 & 0.475 & 0.417 & 0.225 & 0.083 & 1483 & 11 \\
 &  & Gemini-2.5-Flash & 0.253 & 0.362 & 0.393 & 0.175 & 0.083 & 682 & 7 \\
 &  & GPT-4o & 0.046 & 0.037 & 0.048 & 0.100 & 0.000 & 105 & 3 \\
 & PaperCoder & Claude-Sonnet-4.6 & 0.289 & 0.375 & 0.357 & 0.175 & 0.250 & 4606 & 20 \\
 &  & DeepSeek-V4-Pro & 0.338 & 0.512 & 0.548 & 0.125 & 0.167 & 4329 & 8 \\
 &  & Gemini-2.5-Flash & 0.230 & 0.388 & 0.191 & 0.175 & 0.167 & 1203 & 10 \\
 &  & GPT-4o & 0.171 & 0.237 & 0.131 & 0.150 & 0.167 & 731 & 6 \\
 & OpenHands & Claude-Sonnet-4.6 & 0.333 & 0.450 & 0.405 & 0.225 & 0.250 & 1584 & 9 \\
 &  & DeepSeek-V4-Pro & 0.296 & 0.388 & 0.429 & 0.200 & 0.167 & 1400 & 21 \\
 &  & Gemini-2.5-Flash & 0.327 & 0.500 & 0.524 & 0.200 & 0.083 & 475 & 6 \\
 &  & GPT-4o & 0.028 & 0.037 & 0.024 & 0.050 & 0.000 & 324 & 6 \\
\addlinespace
sam2 & BasicAgent & Claude-Sonnet-4.6 & 0.201 & 0.259 & 0.217 & 0.179 & 0.150 & 1206 & 16 \\
 &  & DeepSeek-V4-Pro & 0.306 & 0.429 & 0.293 & 0.202 & 0.300 & 1403 & 22 \\
 &  & Gemini-2.5-Flash & 0.069 & 0.080 & 0.098 & 0.048 & 0.050 & 2737 & 5 \\
 &  & GPT-4o & 0.035 & 0.009 & 0.011 & 0.071 & 0.050 & 110 & 3 \\
 & PaperCoder & Claude-Sonnet-4.6 & 0.215 & 0.375 & 0.196 & 0.191 & 0.100 & 6900 & 28 \\
 &  & DeepSeek-V4-Pro & 0.220 & 0.277 & 0.272 & 0.131 & 0.200 & 6914 & 16 \\
 &  & Gemini-2.5-Flash & 0.202 & 0.223 & 0.228 & 0.155 & 0.200 & 2840 & 20 \\
 &  & GPT-4o & 0.115 & 0.071 & 0.130 & 0.107 & 0.150 & 820 & 6 \\
 & OpenHands & Claude-Sonnet-4.6 & 0.243 & 0.277 & 0.304 & 0.191 & 0.200 & 1585 & 18 \\
 &  & DeepSeek-V4-Pro & 0.219 & 0.259 & 0.250 & 0.167 & 0.200 & 1773 & 14 \\
 &  & Gemini-2.5-Flash & 0.024 & 0.000 & 0.033 & 0.012 & 0.050 & 1541 & 3 \\
 &  & GPT-4o & 0.046 & 0.054 & 0.043 & 0.036 & 0.050 & 621 & 3 \\
\addlinespace
sc-fno & BasicAgent & Claude-Sonnet-4.6 & 0.409 & 0.542 & 0.385 & 0.375 & 0.333 & 1200 & 17 \\
 &  & DeepSeek-V4-Pro & 0.339 & 0.458 & 0.327 & 0.321 & 0.250 & 1255 & 16 \\
 &  & Gemini-2.5-Flash & 0.133 & 0.104 & 0.173 & 0.089 & 0.167 & 682 & 6 \\
 &  & GPT-4o & 0.127 & 0.042 & 0.173 & 0.125 & 0.167 & 188 & 3 \\
 & PaperCoder & Claude-Sonnet-4.6 & 0.310 & 0.354 & 0.308 & 0.161 & 0.417 & 4636 & 23 \\
 &  & DeepSeek-V4-Pro & 0.283 & 0.396 & 0.288 & 0.196 & 0.250 & 4987 & 12 \\
 &  & Gemini-2.5-Flash & 0.274 & 0.396 & 0.288 & 0.161 & 0.250 & 2170 & 15 \\
 &  & GPT-4o & 0.268 & 0.396 & 0.211 & 0.214 & 0.250 & 915 & 8 \\
 & OpenHands & Claude-Sonnet-4.6 & 0.299 & 0.396 & 0.269 & 0.196 & 0.333 & 4225 & 17 \\
 &  & DeepSeek-V4-Pro & 0.305 & 0.312 & 0.308 & 0.268 & 0.333 & 2894 & 19 \\
 &  & Gemini-2.5-Flash & 0.247 & 0.375 & 0.231 & 0.214 & 0.167 & 2338 & 13 \\
 &  & GPT-4o & 0.090 & 0.062 & 0.096 & 0.036 & 0.167 & 380 & 5 \\
\addlinespace
score & BasicAgent & Claude-Sonnet-4.6 & 0.272 & 0.333 & 0.365 & 0.222 & 0.167 & 974 & 14 \\
 &  & DeepSeek-V4-Pro & 0.323 & 0.500 & 0.404 & 0.222 & 0.167 & 1077 & 17 \\
 &  & Gemini-2.5-Flash & 0.196 & 0.250 & 0.250 & 0.056 & 0.229 & 374 & 7 \\
 &  & GPT-4o & 0.138 & 0.125 & 0.211 & 0.028 & 0.188 & 117 & 5 \\
 & PaperCoder & Claude-Sonnet-4.6 & 0.342 & 0.542 & 0.481 & 0.139 & 0.208 & 3223 & 17 \\
 &  & DeepSeek-V4-Pro & 0.223 & 0.333 & 0.308 & 0.083 & 0.167 & 2717 & 7 \\
 &  & Gemini-2.5-Flash & 0.203 & 0.292 & 0.231 & 0.083 & 0.208 & 1408 & 12 \\
 &  & GPT-4o & 0.231 & 0.333 & 0.231 & 0.111 & 0.250 & 1057 & 8 \\
 & OpenHands & Claude-Sonnet-4.6 & 0.281 & 0.375 & 0.269 & 0.250 & 0.229 & 1024 & 11 \\
 &  & DeepSeek-V4-Pro & 0.270 & 0.333 & 0.269 & 0.250 & 0.229 & 1408 & 10 \\
 &  & Gemini-2.5-Flash & 0.246 & 0.417 & 0.211 & 0.167 & 0.188 & 604 & 7 \\
 &  & GPT-4o & 0.128 & 0.125 & 0.192 & 0.111 & 0.083 & 409 & 7 \\
\addlinespace
uno & BasicAgent & Claude-Sonnet-4.6 & 0.120 & 0.250 & 0.147 & 0.083 & 0.000 & 977 & 24 \\
 &  & DeepSeek-V4-Pro & 0.226 & 0.208 & 0.162 & 0.083 & 0.450 & 1396 & 17 \\
 &  & Gemini-2.5-Flash & 0.235 & 0.208 & 0.265 & 0.167 & 0.300 & 2263 & 12 \\
 &  & GPT-4o & 0.067 & 0.042 & 0.176 & 0.000 & 0.050 & 81 & 1 \\
 & PaperCoder & Claude-Sonnet-4.6 & 0.226 & 0.458 & 0.162 & 0.083 & 0.200 & 4614 & 20 \\
 &  & DeepSeek-V4-Pro & 0.214 & 0.375 & 0.073 & 0.208 & 0.200 & 3015 & 6 \\
 &  & Gemini-2.5-Flash & 0.250 & 0.292 & 0.250 & 0.208 & 0.250 & 1491 & 14 \\
 &  & GPT-4o & 0.200 & 0.292 & 0.176 & 0.083 & 0.250 & 835 & 7 \\
 & OpenHands & Claude-Sonnet-4.6 & 0.239 & 0.292 & 0.206 & 0.208 & 0.250 & 882 & 6 \\
 &  & DeepSeek-V4-Pro & 0.257 & 0.292 & 0.235 & 0.250 & 0.250 & 1712 & 6 \\
 &  & Gemini-2.5-Flash & 0.236 & 0.208 & 0.235 & 0.250 & 0.250 & 424 & 6 \\
 &  & GPT-4o & 0.135 & 0.167 & 0.191 & 0.083 & 0.100 & 222 & 5 \\
\addlinespace
voting-lb & BasicAgent & Claude-Sonnet-4.6 & 0.325 & 0.308 & 0.325 & 0.417 & 0.250 & 1001 & 12 \\
 &  & DeepSeek-V4-Pro & 0.319 & 0.404 & 0.400 & 0.222 & 0.250 & 1041 & 10 \\
 &  & Gemini-2.5-Flash & 0.189 & 0.231 & 0.150 & 0.167 & 0.208 & 570 & 3 \\
 &  & GPT-4o & 0.118 & 0.058 & 0.200 & 0.028 & 0.188 & 147 & 3 \\
 & PaperCoder & Claude-Sonnet-4.6 & 0.226 & 0.192 & 0.300 & 0.139 & 0.271 & 3473 & 19 \\
 &  & DeepSeek-V4-Pro & 0.255 & 0.288 & 0.300 & 0.139 & 0.292 & 2953 & 10 \\
 &  & Gemini-2.5-Flash & 0.317 & 0.346 & 0.400 & 0.250 & 0.271 & 964 & 8 \\
 &  & GPT-4o & 0.151 & 0.135 & 0.150 & 0.111 & 0.208 & 747 & 6 \\
 & OpenHands & Claude-Sonnet-4.6 & 0.359 & 0.442 & 0.300 & 0.361 & 0.333 & 1054 & 9 \\
 &  & DeepSeek-V4-Pro & 0.320 & 0.481 & 0.325 & 0.222 & 0.250 & 994 & 10 \\
 &  & Gemini-2.5-Flash & 0.290 & 0.269 & 0.400 & 0.222 & 0.271 & 397 & 5 \\
 &  & GPT-4o & 0.068 & 0.058 & 0.125 & 0.028 & 0.062 & 349 & 6 \\
\addlinespace
wdno & BasicAgent & Claude-Sonnet-4.6 & 0.204 & 0.250 & 0.289 & 0.154 & 0.125 & 1093 & 20 \\
 &  & DeepSeek-V4-Pro & 0.152 & 0.283 & 0.117 & 0.125 & 0.083 & 1019 & 14 \\
 &  & Gemini-2.5-Flash & 0.120 & 0.050 & 0.148 & 0.115 & 0.167 & 1379 & 7 \\
 &  & GPT-4o & 0.073 & 0.008 & 0.078 & 0.038 & 0.167 & 197 & 4 \\
 & PaperCoder & Claude-Sonnet-4.6 & 0.295 & 0.458 & 0.281 & 0.192 & 0.250 & 3782 & 19 \\
 &  & DeepSeek-V4-Pro & 0.174 & 0.175 & 0.180 & 0.135 & 0.208 & 5471 & 12 \\
 &  & Gemini-2.5-Flash & 0.203 & 0.225 & 0.242 & 0.135 & 0.208 & 1963 & 10 \\
 &  & GPT-4o & 0.172 & 0.158 & 0.219 & 0.144 & 0.167 & 918 & 8 \\
 & OpenHands & Claude-Sonnet-4.6 & 0.291 & 0.342 & 0.328 & 0.202 & 0.292 & 1472 & 25 \\
 &  & DeepSeek-V4-Pro & 0.255 & 0.300 & 0.305 & 0.164 & 0.250 & 1625 & 20 \\
 &  & Gemini-2.5-Flash & 0.202 & 0.258 & 0.188 & 0.154 & 0.208 & 2000 & 6 \\
 &  & GPT-4o & 0.086 & 0.067 & 0.062 & 0.048 & 0.167 & 362 & 4 \\
\addlinespace
\bottomrule

\end{longtable}}

\endgroup

\end{document}